\documentclass[aps,10pt,prr,twocolumn,superscriptaddress,floatfix,nofootinbib,longbibliography]{revtex4-2}

\usepackage{graphicx,amsfonts,amssymb,amsmath,hyperref}
\usepackage[ruled,vlined]{algorithm2e}
\usepackage{pgfplots}
\usepackage{pgf}
\usepackage{tikz}
\usetikzlibrary{positioning,arrows.meta}
\usepackage{lmodern}
\usepackage{import}
\usepackage{xr}
\usepackage{enumitem}
\usepackage{tablefootnote}
\usepackage{soul}
\usepackage[normalem]{ulem}
\usepackage{multirow}
\usepackage{booktabs}

\newif\ifhyper
\hypertrue
\ifhyper
\hypersetup{
   citecolor = {red},
   colorlinks = {true},
   urlcolor = {blue}
}
\fi

\newcommand{\beq}{\begin{equation}}
\newcommand{\eeq}{\end{equation}}
\newcommand{\beqa}{\begin{eqnarray}}
\newcommand{\eeqa}{\end{eqnarray}}

\pgfplotsset{compat=1.18}

\usepackage[caption=false]{subfig}
\usepackage{ragged2e}
\usepackage{etoolbox}

\begin{document}

\title{Weight Clustering Reveals the Importance of Relative Rank in Large Language Models}

\author{Borja Aizpurua}
\affiliation{Multiverse Computing, San Sebasti\'{a}n, Spain}
\affiliation{Department of Basic Sciences, Tecnun - University of Navarra, San Sebasti\'an, Spain}

\author{Sukhbinder Singh}
\affiliation{Multiverse Computing, Toronto, Ontario, Canada}

\author{Rom\'{a}n Or\'{u}s}
\affiliation{Multiverse Computing, San Sebasti\'{a}n, Spain}
\affiliation{Donostia International Physics Center, San Sebasti\'an, Spain}
\affiliation{Ikerbasque Foundation for Science, Bilbao, Spain}

\begin{abstract}
Large language models (LLMs) contain billions of parameters, but many exact weight values are not essential. We apply scalar $K$-means clustering to pretrained models, replacing each weight matrix with only 16--64 shared values. For Llama~3.1-8B-Instruct and SmolLM2-135M, this preserves strong accuracy without retraining, providing a simple way to compress LLMs on disk. Fine-tuning only the cluster centroids further recovers 30--40\% of the remaining accuracy loss. Using clustering as an analysis tool, we then keep the cluster assignments fixed and perturb only the shared values, to study which of their properties matter most. We find that preserving their \emph{relative rank} (which connections are stronger or weaker than others) is more important than preserving their exact magnitudes. Breaking this ordering greatly increases perplexity, while changing the values but keeping the ordering causes little loss in many middle and late layers. However, rank preservation alone is not sufficient. Model performance also depends on maintaining the variance of the reconstructed weights, while early layers remain more sensitive; perturbing the cluster assignments themselves shows that their spatial geometry is also load-bearing. We observe the same behavior across four LLMs from two architecture families.
These results suggest that the ordering of weights, together with their overall variance, plays a central role in LLM compression and robustness.
\end{abstract}

\maketitle

\section{Introduction}
\label{sec:intro}

Large language models (LLMs) continue to improve rapidly across a wide range of language tasks~\citep{brown2020gpt3,touvron2023llama}. Yet we still do not understand, in simple terms, how they work. A central question is: how tightly are a model's capabilities tied to the exact numerical values of its weights?

Modern compression methods already suggest the answer: not very tightly. Techniques such as pruning and quantization substantially modify model weights while largely preserving performance. Pruning sets a subset of weights to zero in an already trained model. Quantization reduces numerical precision, for example by converting weights from \texttt{float16} to \texttt{int8}. In many cases, these transformations cause little to no drop in accuracy. Sometimes light retraining after compression---or modest adjustments during pretraining---helps recover any small performance loss. These results indicate that precise weight values may be far less important than commonly assumed.
At the same time, other findings point in the opposite direction: quantization methods must handle outlier channels at full precision to avoid collapse~\citep{dettmers2022gptint}, and recent work has identified individual ``superweights'' whose removal alone destroys model function~\citep{superweight2024}. Taken together, these observations suggest that \emph{some} weight structure is essential --- but which?

\begin{figure}
    \centering
    \includegraphics[width=1\linewidth]{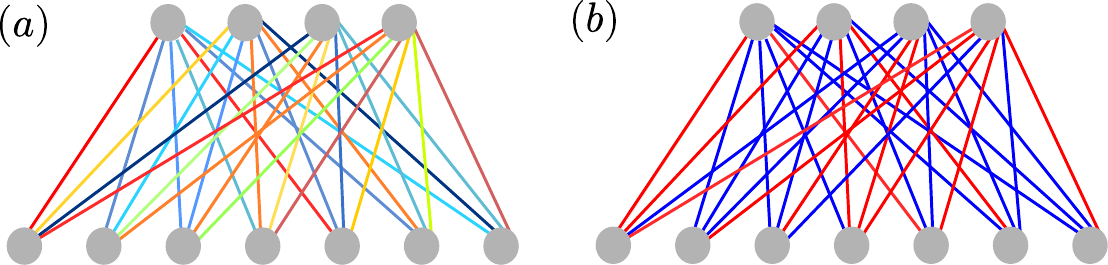}
    \caption{(a) A neural network layer showing connections between input neurons (top) and output neurons (bottom). Edge colors indicate the magnitude of the weights. (b) A clustered version of the same weight matrix with $K=2$ clusters. In this paper, we demonstrate that when the clustered matrix remains performant, model accuracy depends chiefly on the relative clustering of the weights (here, cold vs. hot) rather than on their exact values. The absolute scale of each cluster can be changed by an affine transformation that preserves the overall weight distribution without substantially degrading model performance.}
    \label{fig:cluster}
\end{figure}

An even more aggressive form of compression is possible using a simple method: \emph{weight clustering} with scalar $K$-means~\citep{gong2014compressing,han2015deep}. We group all entries of a pretrained weight matrix into $K$ clusters and replace each value with its cluster representative, see Fig.~\ref{fig:cluster}. For example, with $K = 32$, a matrix containing 58 million distinct values is reduced to just 32 shared values---a reduction of nearly six orders of magnitude. Remarkably, the model often continues to perform well despite this extreme averaging, and it does so without any retraining.

\begin{figure*}[!t]
\centering
\includegraphics[width=0.9\textwidth]{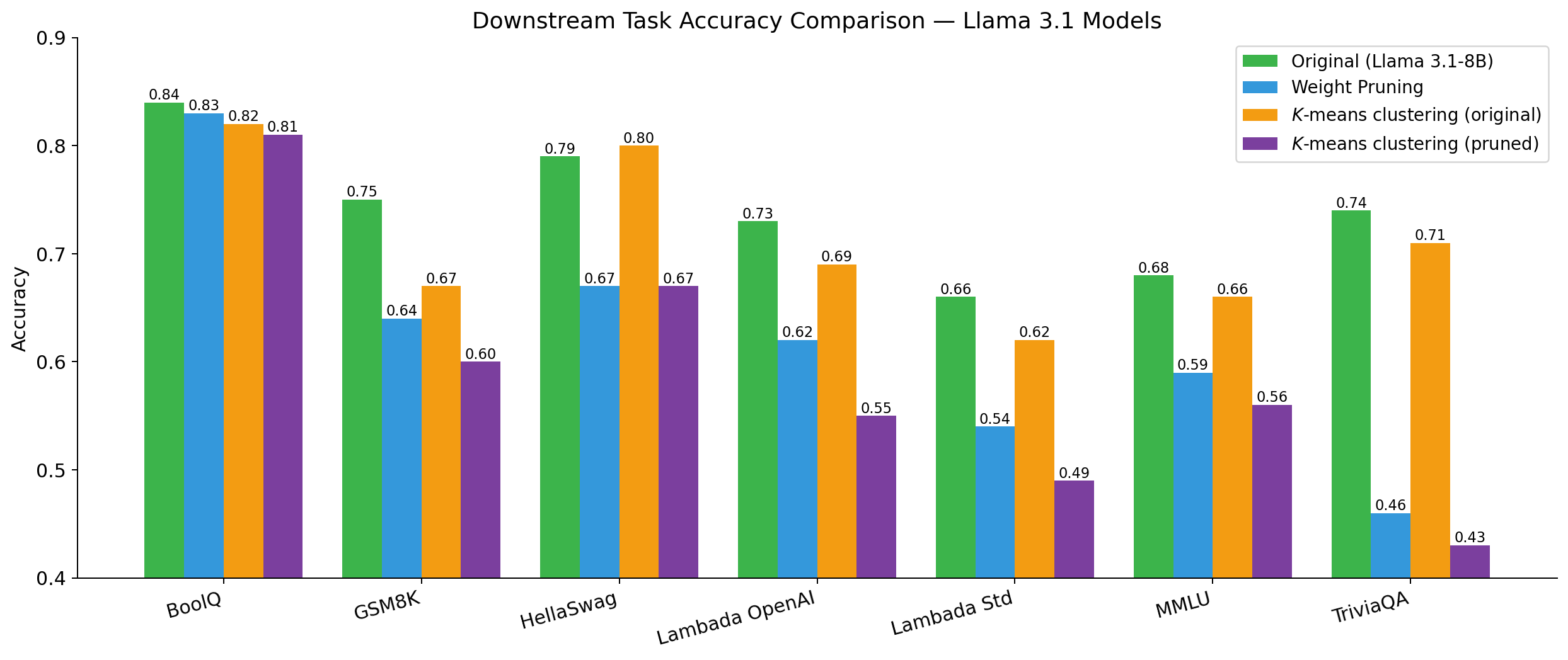}
\caption{Downstream task accuracy comparison for clustered versions of Llama~3.1-8B-Instruct and a compressed Llama 3B model. We see that weight clustering largely preserves accuracy across all tasks without requiring any fine-tuning. Note that the y-axis starts at 0.40, not 0.}
\label{fig:llama8b_downstream}
\end{figure*}

In this paper, we take a closer look at what structure truly matters inside weight-clustered models. We systematically modify the cluster means while keeping the cluster assignments fixed. Our results show that the exact numerical values of the cluster means are largely not what preserve accuracy; what matters instead is their \emph{relative rank}, on equal footing with the overall variance of the reconstructed weights. Across all layer types we tested, rank-preserving changes consistently cause far less damage than rank-breaking ones. Rank preservation alone, however, is not sufficient. Early layers are particularly sensitive to shifts in scale and offset, and extreme nonlinear rescalings can degrade performance even when rank order is maintained. We therefore state the claim conditionally: with the cluster-assignment matrix $L$ held fixed, centroid rank and the reconstructed-weight variance are the two factors that most strongly separate safe modifications from destructive ones, while scale drift can be substantially delayed by affine correction and rank distortion has proven uncorrectable once introduced. Two controls make this conditioning explicit. First, a controlled sweep that varies rank preservation and distribution matching independently identifies centroid rank and the reconstructed-weight variance as two co-equal factors; a partial-correlation analysis shows that rank predicts degradation even after distance and the first two moments are held fixed (strongly so on the 8B model), while distance from the original weights alone does not order the damage. Second, perturbing $L$ itself---rather than the centroids---shows that the spatial assignment geometry is also load-bearing, so the rank effect is established \emph{conditional on} a fixed assignment, not as a claim that rank is the only relevant property. The single-layer rank experiments further replicate on a non-Llama family, for four models across two architecture families in total. This finding has a natural theoretical basis: for any linear layer $y = Wx$, an affine weight transform $w \to aw + b$ with $a>0$ produces an output whose ranking across neurons is unchanged, and whose multiplicative scale is absorbed by the subsequent RMSNorm (which divides by the activation RMS); the additive offset, however, is \emph{not} removed by RMSNorm and is only partially diluted by the residual stream, which is the mechanism behind the early-layer sensitivity above.

A practically important finding is that simple weight clustering largely preserves downstream accuracy across a range of tasks (Fig.~\ref{fig:llama8b_downstream}; per-task breakdown in Appendix~\ref{app:pertask}, Table~\ref{tab:pertask}) \emph{without} requiring any post-clustering training. Furthermore, optional fine-tuning (``healing'') of only the cluster mean values recovers an additional 30--40\% of the accuracy gap at minimal cost. Unless explicitly stated, all compression results and experiments in this paper use no post-compression training of any kind.

\textbf{Relation to existing work.}
Weight clustering for neural networks dates back to the ``Deep Compression'' work of \citet{han2015deep} and earlier results by \citet{gong2014compressing}. They showed that simple $K$-means clustering can compress CNN weights by 16--24$\times$ with little loss in accuracy.

Modern LLM compression methods use more advanced techniques, including non-uniform or learned-codebook quantization (e.g., AQLM~\citep{egiazarian2024aqlm}, VPTQ~\citep{vptq2024}, SqueezeLLM~\citep{kim2024squeezellm}, GANQ~\citep{ganq2024}), low matrix or tensor rank decompositions~\citep{rausch2024fermigrad}, and train-time clustering~\citep{edkm2024}. Our contribution is different. We do not propose a new compression method. Instead, we use weight clustering as a tool to study implicit structure in LLM weights. Our goal is to identify which properties of LLM weights are strongly correlated with model performance. Knowing which properties are load-bearing is useful beyond clustering, since it indicates what any codebook-based or monotone compression scheme must preserve; we return to this in Section~\ref{sec:conclusions}.

Standard post-training quantization (PTQ) methods such as GPTQ~\citep{frantar2023gptq}, AWQ~\citep{lin2023awq}, and SmoothQuant~\citep{xiao2023smoothquant} also tend to preserve the relative ordering of weights, although this property is rarely emphasized as central to their performance.
Here, we show, in a simpler setting, that weight ranks are indeed strongly correlated with model accuracy. Weight clustering provides a clean framework for this analysis. Because cluster assignments and their representative values are stored separately, we can modify the representatives while keeping assignments fixed. This separation lets us directly test how changes in relative rank affect performance.

\section{$K$-means clustering of LLM weights}
\label{sec:method}

By applying scalar $K$-means clustering to a weight matrix $W \in \mathbb{R}^{D \times O}$ we can assign all the weights in $W$ to $K$ clusters of distinct values. As a result, we obtain a (clustered) matrix $\hat{W} \in \mathbb{R}^{D \times O}$ given by:
\beq
\hat{W}[d, o] = c_{L[d,o]}, \label{eq:reconstruction}
\eeq
where the \emph{cluster label matrix} $L[d,o] \in \{1,\ldots,K\}^{D\times O}$ records the parent cluster for each individual weight $W[d, o]$, and $c_k $ is the \emph{centroid} (mean) of the $k^{\mbox{\tiny th}}$ cluster.\footnote{In Eq.~\ref{eq:reconstruction}, $[.,.]$ denotes matrix/vector indexing, e.g., $\hat{W}[d, o]$ denotes a weight located at row $d$ and column $o$.}

The clustered weights $\hat{W}$ can be stored compactly on disk: Cluster labels are stored as packed $\lceil\log_2 K\rceil$-bit integers, while the typically short centroid vector adds negligible overhead (comprised of $K$ float16 values). For instance, with $K=32$, a 58.7M-weight MLP projection is represented by 32 shared values --- a $1.8\times10^6\times$ reduction in distinct values (not end-to-end compression ratio) --- achieving $2$--$4\times$ storage reduction when labels are bit-packed and entropy-compressed.

\section{Experiments: Randomizing weight-clustered LLMs}
\label{sec:ablation}

The central question of this paper is: In Eq.~\ref{eq:reconstruction}, if and how can we change the centroid values $c$ inside a clustered model while keeping the cluster labels $L$ of individual weights fixed, without significantly degrading the model? To study this, we performed experiments on four models: (1) Llama 3.1-8B-Instruct, (2) SmolLM2-135M, (3) a compressed Llama 3B variant, and (4) Qwen2.5-7B-Instruct~\citep{qwen2024}, a non-Llama family on which we replicate the single-layer rank experiments (setup and results in Appendix~\ref{app:qwen}). Models 1, 2, and 4 are open-source, while Model 3 is a compressed version of Llama obtained by applying the CompactifAI algorithm~\citep{tomut2024compactifai}. For readability the detailed single-layer perturbation tables below report the three Llama-architecture models (1)--(3) inline, with the corresponding Qwen2.5 results collected in Appendix~\ref{app:qwen}.

The first step in our experiments was to obtain a clustered version of all three models. For each model, we applied clustering to all attention --- namely, the Query ($Q$), Key $(K$), Value ($V$), and Output ($O$) weights --- and MLP projection matrices --- namely, the gate, up, and down projections ---, while embeddings, layer norms, and the LM head are left unchanged. We fixed the number of clusters, $K$, independently for all layers, including possibly a different $K$ for the different components (attention $Q/K/V/O$ projections, MLP gate/up/down projections) within each transformer block, using a sensitivity criterion based on perplexity (a standard measure of text prediction quality; lower is better). Specifically, for any given layer, we chose a $K$ that produced a clustered weight matrix that degraded the model perplexity by at most 0.5.
This allows assigning a larger $K$ to more sensitive layers (the resulting per-layer $K$ distribution is given in Appendix~\ref{app:k_distribution}).

After determining optimal layerwise values of $K$ as described above, we applied $K$-means clustering to the three models (1)--(3) listed previously, without any subsequent fine-tuning. For SmolLM2-135M, weight-clustering all attention and MLP layers reduces $\sim$106M distinct values to $\sim$10.5k centroids in total; for Llama~3.1-8B-Instruct, the total centroid budget across all clustered layers is 8,192. Applied to Llama~3.1-8B-Instruct with the adaptive per-layer $K$ as described above, weight clustering (without fine-tuning) increases WikiText-2 perplexity from 8.64 to 9.08 (internal harness; see Appendix~\ref{app:eval}). Finally, clustering the 3B compressed variant~\citep{tomut2024compactifai} of Llama~3.1-8B-Instruct (PPL~11.88) yields a clustered model with PPL~12.32.

\subsection{Randomizing single layers}
\label{sec:single_layer}

We first stress a two-stage structure that is easy to conflate. Clustering is run \emph{once} (Section~\ref{sec:method}), producing the fixed label matrix $L$ and centroid vector $c$ of Eq.~\ref{eq:reconstruction}. The experiments here then hold $L$ fixed for the \emph{entire} model and modify only the centroid values of a single layer, $c \mapsto c'$, rebuilding $\hat{W} = c'[L]$. The ``randomized'' centroids below are therefore a \emph{post-clustering} intervention, never an alternative clustering: $L$ stays identical to the baseline until we perturb it directly in Section~\ref{sec:assignment_ablation}.

To begin, we randomized the centroid/mean values (stored in vector $c$ in Eq.~\ref{eq:reconstruction}) inside a single selected layer within the clustered baseline LLM to probe for three distinct effects. First, is scrambling cluster rank more degrading than preserving cluster ranks? Second, should we pay attention to the sign of the cluster means, even when preserving the cluster ranks? And third, how does the impact of these perturbations vary with depth of the layer within the model.

All experiments are carried out on three weight-clustered models --- Llama~3.1-8B-Instruct (baseline PPL~9.08), a pre-compressed Llama~3B (PPL~12.32), and SmolLM2-135M (PPL~27.52) --- by randomizing the centroid values in some way and reconstructing modified weight matrices using Eq.~\eqref{eq:reconstruction}. The Llama~3.1-8B-Instruct and SmolLM2-135M models are publicly available, making these results directly replicable.

\begin{table*}[!t]
\centering
\caption{Single-layer centroid replacement (layer~10 gate\_proj, $K{=}32$).
We replace the centroid values in a single MLP projection layer while preserving the cluster assignments throughout the model.
``Rank-preserving'' replacements preserve the rank ordering of the centroid values.
``Random'' replacements substitute centroid values from a distribution with the same mean and variance as the original values, but do not preserve centroid ranks.
``Rand.\ perm.'' applies a random permutation to the centroid values, the most extreme form of rank-breaking.
Rel.\ $L_2$ change measures how much the reconstructed weight matrix changed.
$\mu$ and $\sigma^2$ are the mean and variance of the reconstructed weights.}
\label{tab:centroid_replacement}
\small
\setlength{\tabcolsep}{2.5pt}
\begin{tabular}{@{}l
    cccc
    cccc
    cccc@{}}
\toprule
& \multicolumn{4}{c}{\textbf{Llama 3.1-8B-Instruct}}
& \multicolumn{4}{c}{\textbf{Compressed Llama 3B}}
& \multicolumn{4}{c}{\textbf{SmolLM2-135M}} \\
\cmidrule(lr){2-5} \cmidrule(lr){6-9} \cmidrule(lr){10-13}
\textbf{\shortstack[l]{Replacement\\mode}}
& \textbf{PPL}
& \textbf{\shortstack{$L_2$\\rel.}}
& \shortstack{$\boldsymbol{\mu}$\\{\scriptsize$(\!\times\! 10^{-4})$}}
& \shortstack{$\boldsymbol{\sigma^2}$\\{\scriptsize$(\!\times\! 10^{-4})$}}
& \textbf{PPL}
& \textbf{\shortstack{$L_2$\\rel.}}
& \shortstack{$\boldsymbol{\mu}$\\{\scriptsize$(\!\times\! 10^{-4})$}}
& \shortstack{$\boldsymbol{\sigma^2}$\\{\scriptsize$(\!\times\! 10^{-4})$}}
& \textbf{PPL}
& \textbf{\shortstack{$L_2$\\rel.}}
& \shortstack{$\boldsymbol{\mu}$\\{\scriptsize$(\!\times\! 10^{-4})$}}
& \shortstack{$\boldsymbol{\sigma^2}$\\{\scriptsize$(\!\times\! 10^{-4})$}} \\
\midrule
Baseline (orig.)
  & 9.08  & ---  & $-$0.2  & 2.0
  & 12.32 & ---  & $-$0.3  & 2.5
  & 27.52 & ---  & $-$1.1  & 324.4 \\
Ord-pres.\ (sorted)
  & 9.17  & 0.52 & $-$16.8 & 0.6
  & 12.46 & 0.62 & $-$6.3  & 4.5
  & 28.35 & 0.52 & $-$234.4 & 94.5 \\
Ord-pres.\ (wt.\ freq.)
  & 9.17  & 0.53 & $-$3.0  & 0.8
  & 12.48 & 0.54 & $-$1.8  & 0.6
  & 28.28 & 0.54 & $-$48.6  & 109.8 \\
Ord-pres., sign-scr.
  & 9.13  & 0.51 & $-$73.4 & 2.0
  & 12.33 & 0.31 & $+$16.8 & 2.5
  & 27.53 & 0.09 & $-$157.7 & 324.5 \\
Sign-pres., ord-brk.
  & 9.56  & 1.43 & $-$25.3 & 6.7
  & 13.37 & 1.70 & $-$30.3 & 27.3
  & 36.58 & 1.43 & $-$289.9 & 1{,}069.0 \\
Rand.\ Gauss.\ ($\mu,\sigma$)
  & 9.76  & 1.42 & $+$25.5 & 2.4
  & 20.98 & 3.10 & $+$3.1  & 2.4
  & 43.92 & 1.41 & $+$257.4 & 384.9 \\
Rand.\ perm.\ (full)
  & 20.67 & 2.69 & $+$0.0  & 10.9
  & 460.2 & 1.99 & $-$19.4 & 7.2
  & 3{,}151 & 2.88 & $-$133.3 & 2{,}126.7 \\
\bottomrule
\end{tabular}
\end{table*}

Table~\ref{tab:centroid_replacement} summarises the results of the experiments, which we explain in detail in the following subsections. Unless the layer is listed explicitly, all single-layer experiments are performed on \texttt{model.layers.10.mlp.gate\_proj}.\footnote{We selected this mid-depth layer (layer~10 of 32) as a representative layer where the contrast between rank-preserving and rank-breaking perturbations is clear and not dominated by the extreme sensitivity of early layers. In Sec.~\ref{sec:ablation_multilayer} we probe the impact of layer depth on these results.}

\subsubsection{Scrambling Centroid Ranks}
\label{sec:ablation_monotone}

We randomize the $K=32$ centroids of the above mentioned layer, drawing random values from a Gaussian distribution with the same mean ($\mu$) and variance ($\sigma^2$) as the original clustered weight matrix.

First, we replace original centroids with the drawn random values, which generally scrambles the cluster ranks. Across all three models, this causes substantial degradation: PPL rises by $+7\%$ on Llama-Instruct (9.08 $\to$ 9.76), $+70\%$ on the pre-compressed model (12.32 $\to$ 20.98), and $+60\%$ on SmolLM2 (27.52 $\to$ 43.92).

However, next we \emph{sort} the drawn values in ascending order, and replace the original centroids in order of their value, thus, preserving the ranks of the clusters --- the cluster that originally represented ``strongly positive'' weights is still the most positive cluster, the ``near-zero'' cluster remains the least extreme, and so on. In this case, we find that this rank-preserving randomization increases perplexity by only $+1\%$ to $+3\%$ across all three models (e.g., 9.08 $\to$ 9.17 on Llama-Instruct). Frequency-weighted monotone replacements (spacing centroids in proportion to their cluster sizes) yield similarly small degradations.

We conclude that at least for a mid-network layer (we explore depth dependence in Sec.~\ref{sec:ablation_multilayer}), all three models are largely indifferent to the exact centroid values as long as the corresponding cluster ranks are preserved.

\subsubsection{Scrambling Centroid Signs}
\label{sec:signs}

Next, we probed if the numerical sign of the centroid values impacts the randomization even when the cluster ranks are preserved.

We find that when the randomization preserved the numerical sign of the centroids but otherwise scrambled their ranks, perplexity degraded in all three models (e.g., PPL 9.08 $\to$ 9.56 on Llama-Instruct, 12.32 $\to$ 13.37 on the pre-compressed model, 27.52 $\to$ 36.58 on SmolLM2). However, when the centroid ranks are strictly preserved but some centroids change sign\footnote{This can be achieved by applying an affine shift $w' = w + b$, where $b = -(c_\text{min} + c_\text{max})/2$}, perplexity remained near baseline in all cases (e.g., PPL~9.13 on Llama-Instruct, 12.33 on the pre-compressed model, 27.53 on SmolLM2).

We conclude that the global rank order matters more than sign structure.
The model maintains the ranks of connection strengths --- not merely their sign --- and this is what must be preserved for the model to function correctly.

\subsubsection{Which Rank-Preserving Randomizations Are Safe?}
\label{sec:ablation_sweep}

Rank-breaking is harmful; but is \emph{every} rank-preserving transform safe? We applied a family of strictly monotone maps $f$ to the centroids of \texttt{layers.10.mlp.gate\_proj} ($L$ fixed): linear rescalings $f(w) = aw$ ($a \in \{0.5, 2\}$), saturating maps $f(w) = \tanh(\alpha w)$ ($\alpha \in \{1, 2, 3\}$), and power maps $f(w) = \mathrm{sign}(w)|w|^{\gamma}$ ($\gamma \in \{0.5, 1.5\}$). All preserve centroid rank exactly and differ only in how they space the centroids---and hence in the mean and variance of the reconstructed weights (Table~\ref{tab:monotone_sweep}).

\begin{table*}[t]
\centering
\caption{Monotone centroid transformations on \texttt{layers.10.mlp.gate\_proj} (labels fixed).
All transforms preserve centroid rank order.
$\mu$ and $\sigma^2$ are the mean and variance of the reconstructed weight matrix $\hat{W}$ after each transform.
The mean stays near zero throughout (the centroid distribution is symmetric); variance is the informative predictor of degradation.}
\label{tab:monotone_sweep}
\small
\begin{tabular}{l
    ccc
    ccc
    ccc}
\toprule
& \multicolumn{3}{c}{\textbf{Llama 3.1-8B-Instruct}}
& \multicolumn{3}{c}{\textbf{Compressed Llama 3B}}
& \multicolumn{3}{c}{\textbf{SmolLM2-135M}\,$^\dagger$} \\
\cmidrule(lr){2-4} \cmidrule(lr){5-7} \cmidrule(lr){8-10}
\textbf{Transform}
& \textbf{PPL}
& \shortstack{$\boldsymbol{\mu}$\\{\scriptsize$(\times 10^{-4})$}}
& \shortstack{$\boldsymbol{\sigma^2}$\\{\scriptsize$(\times 10^{-4})$}}
& \textbf{PPL}
& \shortstack{$\boldsymbol{\mu}$\\{\scriptsize$(\times 10^{-4})$}}
& \shortstack{$\boldsymbol{\sigma^2}$\\{\scriptsize$(\times 10^{-4})$}}
& \textbf{PPL}
& \shortstack{$\boldsymbol{\mu}$\\{\scriptsize$(\times 10^{-2})$}}
& \shortstack{$\boldsymbol{\sigma^2}$\\{\scriptsize$(\times 10^{-2})$}} \\
\midrule
Identity
  & 9.08  & $-$0.2  & 2.0
  & 12.32 & $-$0.3  & 2.5
  & 27.52 & $-$1.1  & 3.2 \\
Affine $w \mapsto 0.5w$
  & 9.18  & $-$0.1  & 0.5
  & 12.46 & $-$0.1  & 0.6
  & 28.25 & $-$0.5  & 0.8 \\
Affine $w \mapsto 2.0w$
  & 9.59  & $-$0.3  & 8.2
  & 13.10 & $-$0.6  & 9.9
  & 30.58 & $-$2.2  & 13.0 \\
$\tanh(\alpha w)$, $\alpha = 1$
  & 9.08  & $-$0.2  & 2.0
  & 12.32 & $-$0.3  & 2.5
  & 27.55 & $-$1.1  & 3.0 \\
$\tanh(\alpha w)$, $\alpha = 2$
  & 9.59  & $-$0.3  & 8.1
  & 13.09 & $-$0.6  & 9.9
  & 29.36 & $-$2.3  & 10.3 \\
$\tanh(\alpha w)$, $\alpha = 3$
  & 11.33 & $-$0.5  & 18.3
  & 15.89 & $-$0.8  & 22.2
  & 33.90 & $-$3.3  & 19.0 \\
$\mathrm{sign}(w)|w|^{\gamma}$, $\gamma = 1.5$
  & 9.37  & $-$0.03 & 0.05
  & 12.66 & $-$0.05 & 0.07
  & 28.09 & $-$0.3  & 1.0 \\
$\mathrm{sign}(w)|w|^{\gamma}$, $\gamma = 0.5$
  & \textbf{1{,}159.52} & \textbf{$-$3.6} & \textbf{111.8}
  & \textbf{1{,}510.1}  & \textbf{1.2}    & \textbf{122.3}
  & 31.51 & $-$13.0 & 14.2 \\
\bottomrule
\end{tabular}

\vspace{2pt}
{\footnotesize $^\dagger$\,SmolLM2-135M has larger weight magnitudes; $\mu$ and $\sigma^2$ are scaled by $10^{-2}$ rather than $10^{-4}$.}
\end{table*}

We find that most of these rank-preserving transforms keep PPL near baseline across all three models. The main outlier is power compression with $\gamma=0.5$, which causes catastrophic degradation on Llama-Instruct (PPL~1{,}160) and the pre-compressed model (PPL~1{,}510), though notably only mild degradation on SmolLM2 (PPL~31.51, $1.14\times$ baseline). Tanh $\alpha=3$ also degrades all three models. Power $\gamma=0.5$ expands the tails while Tanh $\alpha=3$ compresses them (large centroids are squeezed toward $\pm 1$); both distort the weight variance (evident from Table~\ref{tab:monotone_sweep}) despite preserving rank order, which we believe drives the degradation.

\begin{figure}[t]
\centering
\includegraphics[width=0.99\columnwidth]{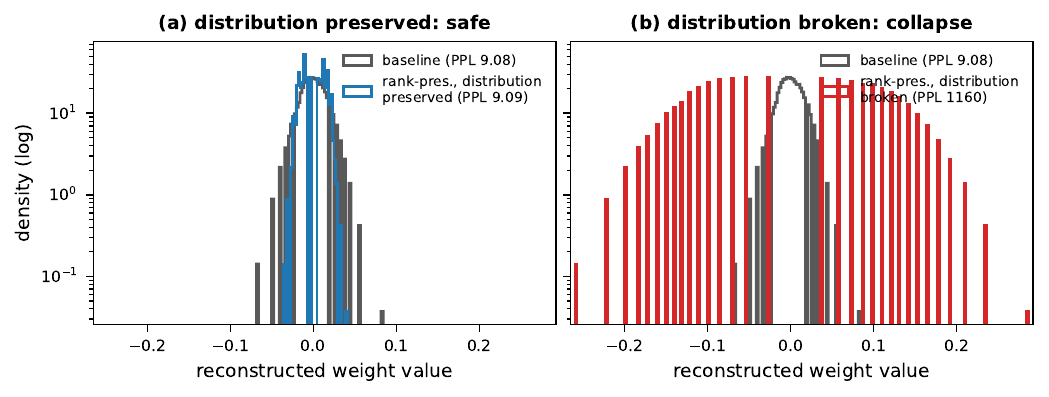}
\caption{Why the weight \emph{distribution}, not just rank, matters (\texttt{layers.10.mlp.gate\_proj}, Llama-3.1-8B-Instruct, $K{=}32$; log-density $y$-axis). Both panels apply the \emph{same} monotone warp $c \mapsto \mathrm{sign}(c)|c|^{0.5}$, so centroid rank is identical; they differ only in the distribution. \textbf{(a)} Renormalized to the original mean and variance (Eqs.~\ref{eq:mean_cond}--\ref{eq:var_cond}), the distribution matches baseline and PPL is unchanged ($9.08\!\to\!9.09$). \textbf{(b)} Without renormalization the variance inflates $55\times$ and PPL collapses to $1{,}160$. With rank held fixed, it is the distribution---chiefly the variance---that separates the safe edit from the destructive one.}
\label{fig:weight_dist}
\end{figure}

A nonlinear monotone map changes the mean, variance, higher moments, and distance from the original weights all at once, so to confirm that the \emph{distribution} is the culprit we hold the warp---and thus the rank---fixed and toggle only the distribution (Fig.~\ref{fig:weight_dist}): renormalizing the $\gamma{=}0.5$ warp back to the original mean and variance restores baseline perplexity ($9.09$), whereas the same warp left un-normalized inflates the variance $55\times$ and collapses the model to PPL~$1{,}160$---with identical centroid rank in both cases. This isolates the weight variance as the operative property (we control for distance explicitly in Section~\ref{sec:controlled_grid}).

These results suggest that \emph{preserving ordering alone is not sufficient} --- the overall weight distribution must also be preserved. Therefore, a broad class of ``safe'' transformations of the cluster means, namely, maps $f: c_k \mapsto f(c_k)$ are those that preserve both ordering and the mean ($\mu$) and variance ($\sigma^2$) of the reconstructed weight distribution:
\begin{align}
    \sum_k p_k f(c_k) &= \sum_k p_k c_k \quad \text{(preserve $\mu$)}, \label{eq:mean_cond} \\
    \sum_k p_k [f(c_k) - \bar{c}]^2 &= \sum_k p_k [c_k - \bar{c}]^2 \quad \text{(preserve $\sigma^2$)}. \label{eq:var_cond}
\end{align}
Affine maps
\beq
f(w) = a w + b, \quad a > 0 \label{eq:affine}
\eeq
always preserve rank. Applied to a set of replacement centroids $c'_k$ with mean $\mu'$ and variance $\sigma'^2$ (under the same weights $p_k$), the choice $a = \sigma/\sigma'$, $b = \mu - a\mu'$ also satisfies Eqs.~\ref{eq:mean_cond}--\ref{eq:var_cond}. (Applied to the original centroids, the only affine map satisfying both conditions is the identity.)
To gain some intuition as to why such affine transformation can preserve performance in some situations, consider any linear layer $W$ inside the network. If every entry of $W$ undergoes the affine transform $w \to aw + b$ with $a > 0$ the output activations $y = Wx$ of the layer transform as
\beq
y_o \;\to\; a\,y_o \;+\; b\!\sum_{d} x_d\,,
\label{eq:affine_output}
\eeq
where the offset $b\sum_{d} x_d$ is identical for every output neuron~$o$. Because the offset is the same across all outputs, the ranks of the activations $y_1, y_2, \ldots, y_O$ are preserved for any $a > 0$. In other words, the ordering of neurons --- those that fire most strongly, those near zero, and those most negative --- remains unchanged. Subsequent monotone nonlinearities, including SiLU in SwiGLU blocks, therefore preserve this ordering as well, while propagating the scale and offset through the network until encountering a normalization layer. Llama~3.1 and SmolLM2 both use \textbf{RMSNorm} rather than LayerNorm, and the distinction matters here. RMSNorm divides the activation vector by its root-mean-square but, unlike LayerNorm, does \emph{not} subtract the mean. Consequently the two parts of the affine map are absorbed asymmetrically: the multiplicative factor $a$ rescales the RMS and is therefore absorbed (post-norm activations are nearly invariant under $w\mapsto aw$), whereas the additive offset $b\sum_d x_d$ is a constant shift that RMSNorm does \emph{not} remove and that instead propagates, only partially diluted, through the residual stream. A single small affine map is thus a near \emph{gauge symmetry} only through its multiplicative channel. We verified this directly (Appendix~\ref{app:activations}): applying a pure scaling $w\mapsto 0.5w$ to one mid-layer leaves the following block's post-RMSNorm activations at cosine similarity ${\ge}0.98$ with the untransformed model, and a small centred shift (the order-preserving sign-scramble of Section~\ref{sec:signs}) leaves them comparably close ($\ge 0.98$). A large offset of several weight standard deviations, by contrast, drops the similarity substantially (to $0.81$--$0.88$), and rank-breaking replacements also reduce it (to $0.87$--$0.94$), whereas every rank-preserving, distribution-preserving edit stays near $0.98$--$0.99$.

The residual stream thus breaks this symmetry only partially: because each block adds its output to the residual, the affine change is diluted but not fully cancelled, which is why early layers --- sitting at the front of a long residual chain --- are more sensitive (Section~\ref{sec:ablation_multilayer}). Our layer-wise diagnostic confirms this: applying the same offset to layer~0 perturbs the final-layer activations far more than applying it to any later layer (Appendix~\ref{app:activations}). Both halves of this argument---the exact absorption of the multiplicative factor and the non-removal of the offset---are identities rather than approximations, and we state them precisely in Appendix~\ref{app:activations}; they are also the extent of what we derive, since they constrain scale and say nothing about rank (Section~\ref{sec:normal_centroids}).

When the same affine map is applied to \emph{all} layers simultaneously, these residual contributions accumulate and the per-block RMSNorms can no longer absorb the shift exactly. Nevertheless, the global results in Table~\ref{tab:monotone_sweep} ($w\!\mapsto\!0.5w$, $w\!\mapsto\!2w$) show that the residual leakage remains small in practice, consistent with each intermediate RMSNorm partially re-scaling the activations before the next block.

Note that transforms listed in Table \ref{tab:monotone_sweep} that distort higher-order moments (beyond mean and variance) of the weight distribution --- such as the kurtosis expansion from $\gamma = 0.5$ or the tail compression from large-$\alpha$ tanh --- create activation distributions that RMSNorm cannot repair, thus, explaining why variance of the reconstructed weights is the informative predictor of degradation in Table~\ref{tab:monotone_sweep}. We make this quantitative in the controlled sweep of Section~\ref{sec:controlled_grid}, where re-normalizing a rank-preserving transform back to the original mean and variance (e.g.\ $\mathrm{sign}(w)|w|^{0.5}$ followed by an affine rescaling) restores near-baseline perplexity, isolating distributional drift---not rank---as the cause of the $\gamma{=}0.5$ collapse.

\subsubsection{Dependence of Randomization on Layer Depth}
\label{sec:ablation_multilayer}

Next, we probed the impact of randomizing cluster means inside layers at varying depths in the model. For each model we selected an early, mid, and late MLP projection as well as an attention projection; the exact layer indices differ across models because the architectures have different depths (e.g., layer~26 is late for the compressed Llama~3B but mid-network for the 32-layer Llama-Instruct, where we use layer~30 instead). See Table~\ref{tab:multilayer_replacement} for the specific layers used.

\begin{table}[t]
\centering
\caption{
Single-layer centroid replacement at varying depths.
Each cell perturbs only the named layer; all other layers are unmodified.
Rank-breaking conditions (kp sign/brk ord, gauss rand) are consistently more damaging than order-preserving ones (kp ord/brk sign, gauss mono) at every layer.
}
\label{tab:multilayer_replacement}
\small
\setlength{\tabcolsep}{2.5pt}
\begin{tabular}{@{}l cccc@{}}
\toprule
& \textbf{\shortstack{kp\,sign\\brk\,ord}}
& \textbf{\shortstack{kp\,ord\\brk\,sign}}
& \textbf{\shortstack{gauss\\rand}}
& \textbf{\shortstack{gauss\\mono}} \\
\midrule
\multicolumn{5}{@{}l}{\textit{Llama 3.1-8B-Instruct} (PPL\,=\,9.08)} \\[1pt]
\quad l1\,(early)    & 9.64      & 117.39  & 2{,}260    & 11.30 \\
\quad l10\,(mid)     & 9.56      & 9.13    & 9.76       & 9.17 \\
\quad l30\,(late)    & 13.18     & 9.08    & 47.30      & 10.00 \\
\quad l10\,($q$-proj)& 10.17     & 9.08    & 9.83       & 9.48 \\[3pt]
\multicolumn{5}{@{}l}{\textit{Compressed Llama 3B} (PPL\,=\,12.32)} \\[1pt]
\quad l1\,(early)    & 78{,}847  & 24.39   & 28{,}315   & 102.6 \\
\quad l10\,(mid)     & 13.37     & 12.32   & 13.22      & 12.47 \\
\quad l26\,(late)    & 19.56     & 12.33   & 293.7      & 13.35 \\
\quad l10\,($q$-proj)& 15.69     & 12.37   & 13.53      & 13.01 \\[3pt]
\multicolumn{5}{@{}l}{\textit{SmolLM2-135M} (PPL\,=\,27.52)} \\[1pt]
\quad l1\,(early)    & 36.58     & 27.72   & 42.60      & 28.49 \\
\quad l10\,(mid)     & 36.58     & 27.53   & 43.92      & 28.35 \\
\quad l28\,(late)    & 665{,}943 & 28.96   & 69.98      & 39.40 \\
\quad l10\,($q$-proj)& 32.74     & 27.58   & 31.24      & 28.27 \\
\bottomrule
\end{tabular}
\end{table}

The results are listed in Table~\ref{tab:multilayer_replacement}.
First, the qualitative pattern is consistent across all four layers and all three models: in every row, the two rank-breaking columns are more damaging than the two rank-preserving columns.
Thus, this pattern is not an artefact of a single layer or a single model. Because our evidence so far is concentrated in MLP projections, we additionally applied the four perturbation modes to all four attention projections ($Q$, $K$, $V$, $O$) as well as the three MLP projections (gate, up, down) at a common mid layer; the rank-breaking-$>$rank-preserving ordering holds for every projection type (Appendix~\ref{app:coverage}, Table~\ref{tab:coverage}). We use \texttt{mlp.gate\_proj} as the representative single-layer probe because the contrast there is large and not dominated by early-layer sensitivity, but the effect is not specific to it.

Second, while rank preservation is necessary it is not always sufficient to preserve clustering accuracy. For the early layer (layer~1), even the rank-preserving randomization gives a considerable hit to perplexity. In the Instruct model, the sign-scramble (which uses an affine shift) pushes PPL to 117.39, indicating the sign ladder is load-bearing at early layers. In the pre-compressed model, the sorted-Gaussian replacement gives PPL~102.6.
This is a cumulative-drift mechanism, as we explore further in Section~\ref{sec:drift}: early layers sit at the front of the propagation chain, so any scale or offset shift amplifies across the remaining depth.
By contrast, the mid-network and late layers tolerate order-preserving distortions gracefully (PPL within $1.1\times$ baseline).

Third, rank-breaking at early layers is catastrophic: in the pre-compressed model, PPL reaches the tens of thousands; in the Instruct model, Gaussian random replacement at layer~1 collapses to 2{,}260; and in SmolLM2, order-breaking at the late layer (L28) causes an extreme collapse to 665{,}943.

Together, these observations confirm the main thesis while sharply qualifying it: rank-breaking is the main failure mode at every depth; rank-preserving replacements can still hurt via cumulative scale drift, and the penalty grows as the randomized layer sits earlier in the network.

\subsection{Disentangling rank from scale, variance, and distance}
\label{sec:controlled_grid}

A natural objection to the experiments above is that each perturbation changes several quantities at once---rank order, mean, variance, higher moments, and the overall distance from the original weights---so they do not cleanly identify rank order as the operative variable. To separate these factors we ran a controlled sweep on \texttt{layers.10.mlp.gate\_proj} and \texttt{layers.10.self\_attn.q\_proj} in which two factors are varied independently: whether the centroid map is \emph{rank-preserving} (monotone) or \emph{rank-breaking}, and whether the reconstructed weight distribution is \emph{matched} (mean and variance held to the original) or \emph{free}. The sweep spans affine, power, and tanh transforms (rank-preserving), partial and full centroid permutations (rank-breaking but histogram-exact), and Gaussian replacements (rank-breaking, free). For every perturbation we record the relative-$L_2$ distance of the reconstructed weights, the mean and variance drift, and the centroid rank correlation, all computed analytically, alongside the resulting perplexity.

Three contrasts in this sweep separate the variables, and together they refine---rather than simply confirm---the headline claim. \emph{(i) Rank, with the distribution held fixed.} Sorted vs.\ unsorted Gaussian replacement share the same target mean and variance and differ only in whether centroid rank is preserved; the sorted (rank-preserving) version is less damaging at every projection in all three models (Table~\ref{tab:centroid_replacement} and Appendix~\ref{app:coverage}), isolating rank with the distribution controlled. \emph{(ii) Distribution, with rank held fixed.} Conversely, a rank-preserving transform that distorts the variance is harmful---$\mathrm{sign}(w)|w|^{0.5}$ collapses Llama-Instruct ($9.08\!\to\!1159$)---yet the \emph{same} transform re-normalized to the original mean and variance returns to baseline (Llama-Instruct $9.08\!\to\!9.09$; SmolLM2 $31.5\!\to\!27.8$), isolating the distribution. \emph{(iii) Partial-correlation regression.} A standardized regression of $\log(\mathrm{PPL}/\mathrm{PPL}_0)$ on centroid rank correlation, $|\Delta\sigma^2|$, $|\Delta\mu|$, and relative-$L_2$ distance over the full sweep (Table~\ref{tab:controlled_grid}) shows that \emph{rank and variance are the two quantities that remain predictive once distance is controlled}: on Llama \texttt{gate\_proj}, rank correlation has partial correlation $-0.56$ and $|\Delta\sigma^2|$ has $+0.97$, while distance itself does not predict damage in the direction a distance-based account would require (its partial correlation is $-0.01$ on SmolLM2 and $-0.80$ on Llama, i.e.\ not positive). Crucially, distance alone does \emph{not} order the damage---a far rank-\emph{breaking} permutation (rel-$L_2{=}2.3$) leaves Llama at PPL~$12.7$, whereas the far, variance-distorting but rank-\emph{preserving} $\mathrm{sign}(w)|w|^{0.5}$ (rel-$L_2{=}6.5$) reaches $1159$. We therefore state a refined conclusion: centroid rank and the reconstructed-weight variance are two distinct, co-equal factors (the safe class of Eqs.~\ref{eq:mean_cond}--\ref{eq:var_cond}), so rank preservation is necessary but not sufficient---what makes a rank-preserving edit safe is that it also preserves the weight distribution. (On the small model the regression is dominated by mean drift, $|\Delta\mu|$ partial correlation $+0.62$; the SmolLM2 rank effect appears most cleanly in contrast~(i) and in the assignment ablation of Section~\ref{sec:assignment_ablation}, and attention \texttt{q\_proj} is too perturbation-robust---all swept points within ${\sim}1.3\times$ baseline---for its regression to be informative.)

\begin{table}[t]
\centering
\caption{Standardized partial correlations of $\log(\mathrm{PPL}/\mathrm{PPL}_0)$ with each quantity over the controlled sweep on \texttt{layers.10.mlp.gate\_proj} ($n{=}28$ perturbations per model). A large-magnitude value means the quantity predicts degradation \emph{after} the other three are held fixed. On the larger model both centroid rank (negative: more rank preservation $\Rightarrow$ less damage) and variance drift (positive) are strong independent predictors, while distance from the original weights is not.}
\label{tab:controlled_grid}
\small
\setlength{\tabcolsep}{5pt}
\begin{tabular}{l cc}
\toprule
\textbf{Predictor} & \shortstack{\textbf{SmolLM2}\\\textbf{135M}} & \shortstack{\textbf{Llama 3.1}\\\textbf{8B-Inst.}} \\
\midrule
Rank corr.                  & $-0.06$ & $\mathbf{-0.56}$ \\
$|\Delta\sigma^2|$ (var.) & $+0.17$ & $\mathbf{+0.97}$ \\
$|\Delta\mu|$ (mean)         & $\mathbf{+0.62}$ & $+0.47$ \\
rel-$L_2$ (dist.)           & $-0.01$ & $-0.80$ \\
\bottomrule
\end{tabular}
\end{table}

\subsection{Perturbing the cluster assignments $L$}
\label{sec:assignment_ablation}

All experiments above keep the cluster-label matrix $L$ fixed and modify only the centroid values $c$. This is what lets us attribute effects to the centroids, but it leaves open whether the spatial assignment geometry of $L$---which weight is assigned to which cluster---is itself the main carrier of model function. We therefore ran the complementary study: keep the centroid \emph{values} fixed and perturb $L$. We use two families. \emph{Position shuffles} permute which positions receive which value, and so preserve the global value histogram (hence mean, variance, and all higher moments) exactly while destroying spatial structure at different scales: within each row, within each column, within contiguous windows, and globally. \emph{Rank-bounded relabels} keep $L$ fixed but swap which centroid a label points to, restricted either to rank-adjacent clusters (a small value move) or to rank-distant clusters (a large, rank-breaking move).

The results (Table~\ref{tab:assignment}) give a nuanced but consistent picture. Destroying the spatial assignment geometry, even though the value distribution is left untouched, degrades the model in all cases, but to a model-dependent degree: a global position reshuffle of a single mid layer raises SmolLM2 perplexity from $27.5$ to $43.0$ (comparable to a rank-breaking centroid replacement at the same layer, $43.9$), while the larger models are markedly more robust to scrambling one layer's geometry (Llama-Instruct $9.08\!\to\!9.8$; compressed 3B $12.3\!\to\!13.4$). So assignment geometry is load-bearing---most visibly in the smallest model---and the rank claim must be read as conditional on it. The rank-bounded relabels, which hold $L$ fixed, isolate rank cleanly and behave consistently across all models, including the non-Llama Qwen2.5-7B-Instruct (Appendix~\ref{app:qwen}): reassigning weights among rank-\emph{adjacent} clusters is nearly free (all within $0.01\times$ baseline), whereas reassigning the same number among rank-\emph{distant} clusters is severely damaging (SmolLM2 $27.5\!\to\!294$, compressed 3B $12.3\!\to\!718$, Llama-Instruct $9.08\!\to\!28.1$, Qwen2.5-7B $11.8\!\to\!19.8$). It is therefore the \emph{rank distance} of a reassignment, not the act of reassignment, that determines the damage.

\begin{table*}[t]
\centering
\caption{Cluster-assignment ($L$) perturbations on \texttt{layers.10.mlp.gate\_proj}, with centroid values held fixed. Position shuffles preserve the value histogram exactly (hence $\mu,\sigma^2$, and all moments) and vary only spatial structure; rank-bounded relabels keep $L$ fixed and swap labels among rank-adjacent vs.\ rank-distant clusters. PPL is WikiText-2 (internal harness).}
\label{tab:assignment}
\small
\setlength{\tabcolsep}{4pt}
\begin{tabular}{l ccc c}
\toprule
\textbf{$L$ perturbation} & \textbf{SmolLM2} & \textbf{Llama-Inst.} & \textbf{Comp.\ 3B} & \textbf{Qwen2.5} \\
\midrule
Baseline                       & 27.52 & 9.08 & 12.32 & 11.83 \\
\midrule
Within-row shuffle             & 43.83 & 9.85  & 13.43 & 14.56 \\
Within-column shuffle          & 36.42 & 9.88  & 13.41 & 17.63 \\
Window-64 shuffle              & 42.67 & 9.82  & 13.26 & 14.80 \\
Global reshuffle (freq.\ kept) & 43.02 & 9.82  & 13.37 & 14.69 \\
\midrule
Relabel rank-adjacent          & 27.63 & 9.08  & 12.32 & 11.83 \\
Relabel rank-distant           & 294.3 & 28.14 & 718.0 & 19.78 \\
\bottomrule
\end{tabular}
\end{table*}

\subsection{Randomizing Multiple Layers}
\label{sec:drift}

Generally, we expect that small distortions introduced at each layer can compound --- each layer's output becomes a slightly wrong input for the next, and eventually the signal degrades beyond recovery.
To test how errors propagate, we progressively randomized centroids in a rank-preserving way, block-by-block across all transformer blocks of each model, tracking perplexity after every two blocks (replacing from deepest to shallowest, since early layers are known to be more sensitive).

\begin{figure}[t]
\centering
\begin{tikzpicture}
\begin{axis}[
    width=0.95\columnwidth,
    height=0.65\columnwidth,
    xlabel={Fraction of network replaced (\%)},
    ylabel={PPL\,/\,baseline},
    ymode=log,
    xmin=0, xmax=100,
    ymin=0.8, ymax=1e10,
    grid=major,
    legend pos=north east,
    legend style={font=\scriptsize, cells={anchor=west}}
]
\addplot[blue, thin, mark=none, forget plot] coordinates {
    (0, 1) (6.2, 847235) (12.5, 908651) (18.8, 830030) (25.0, 845333)
    (31.2, 1073654) (37.5, 1343053) (43.8, 1360476) (50.0, 1364897)
    (56.2, 1375787) (62.5, 1350680) (68.8, 1370898) (75.0, 1375033)
    (81.2, 1374175) (87.5, 1377494) (93.8, 1383146) (100, 1384073)
};
\addplot[red, thin, mark=none, forget plot] coordinates {
    (0, 1) (7.1, 981640) (14.3, 1060692) (21.4, 1081713) (28.6, 953621)
    (35.7, 226037) (42.9, 900114) (50.0, 1079258) (57.1, 1079783)
    (64.3, 1079443) (71.4, 1083832) (78.6, 1082614) (85.7, 1074467)
    (92.9, 1041667) (100, 1084738)
};
\addplot[teal, thin, mark=none, forget plot] coordinates {
    (0, 1) (6.7, 2975114155) (13.3, 138884154) (20.0, 115039159)
    (26.7, 1219067642) (33.3, 1110673725) (40.0, 7756753562)
    (46.7, 7744079906) (53.3, 7803683229) (60.0, 7956582720)
    (66.7, 8071601410) (73.3, 8904831584) (80.0, 8580708355)
    (86.7, 8496211132) (93.3, 8273506507) (100, 8180241362)
};
\addplot[blue, thick, mark=*, mark size=1.5pt] coordinates {
    (0, 1.00) (6.2, 1.32) (12.5, 1.25) (18.8, 1.31) (25.0, 1.32)
    (31.2, 1.36) (37.5, 1.41) (43.8, 1.47) (50.0, 1.55)
    (56.2, 1.63) (62.5, 2.89) (68.8, 2.56) (75.0, 2.19)
    (81.2, 2.41) (87.5, 3.01) (93.8, 7.54) (100, 4256)
};
\addlegendentry{Llama-3.1-8B-Inst.}
\addplot[red, thick, dashed, mark=square, mark size=1.5pt] coordinates {
    (0, 1.00) (7.1, 1.14) (14.3, 1.26) (21.4, 1.41) (28.6, 1.45)
    (35.7, 1.48) (42.9, 1.65) (50.0, 1.81) (57.1, 2.28)
    (64.3, 2.80) (71.4, 3.58) (78.6, 3.42) (85.7, 5.11)
    (92.9, 21.02) (100, 14155)
};
\addlegendentry{Compressed Llama-3B}
\addplot[teal, thick, dotted, mark=triangle, mark size=1.5pt] coordinates {
    (0, 1.00) (6.7, 1.71) (13.3, 2.82) (20.0, 3.72) (26.7, 4.39)
    (33.3, 5.05) (40.0, 8.11) (46.7, 11.20) (53.3, 20.33)
    (60.0, 33.40) (66.7, 10554375) (73.3, 96310)
    (80.0, 76944) (86.7, 82426) (93.3, 7127453) (100, 51273468)
};
\addlegendentry{SmolLM2-135M}
\end{axis}
\end{tikzpicture}
\caption{Normalized perplexity (PPL\,/\,baseline) as centroids are progressively replaced with rank-preserving-random values, one transformer block at a time from deepest to shallowest. Thick lines with markers show affine-corrected replacement (mean and variance of the reconstructed weights are preserved); thin lines of matching color show uncorrected replacement. Each curve normalizes by its own baseline (Llama~3.1-8B-Instruct: 9.08; compressed Llama~3B: 12.32; SmolLM2-135M: 27.52).}
\label{fig:cumulative_drift}
\end{figure}
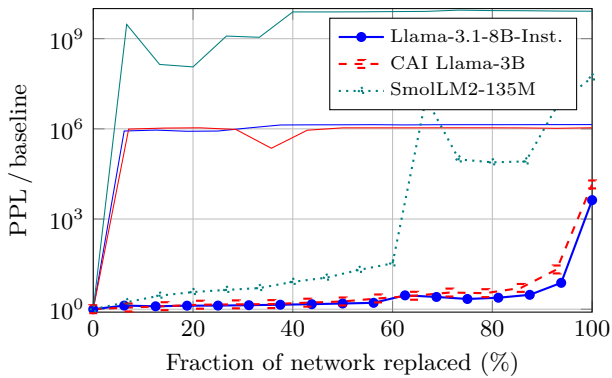

The results are plotted in Fig.~\ref{fig:cumulative_drift}. Without scale correction (thin lines), even two blocks of monotone-random centroid replacement cause immediate catastrophic collapse ($>10^5{\times}$ baseline) in all three models, showing that per-layer scale and offset shifts propagate destructively through the remaining depth.

A key question is whether this collapse is driven specifically by \emph{scale drift} (the reconstructed weight distribution shifting layer by layer) or by some other aspect of the randomized monotone distortion of the centroids.
To isolate this, we repeat the block-by-block randomization while also preserving the weight distribution (mean and variance) by applying, to the randomized replacement values, affine maps (Eq.~\ref{eq:affine}) whose $a$ and $b$ are chosen so that the mean and variance of the reconstructed weights match the original ones (Eqs.~\ref{eq:mean_cond}--\ref{eq:var_cond}).
Under this scale-corrected randomization, the perplexity collapse is substantially delayed and reduced in magnitude (Fig.~\ref{fig:cumulative_drift}). The effect is model-size dependent: Llama~3.1-8B-Instruct stays within $1.6{\times}$ baseline through 56\% of its network; the pre-compressed 3B model stays within $2{\times}$ through ${\sim}50\%$; while SmolLM2-135M degrades faster ($5{\times}$ at one-third of the network) and collapses entirely at ${\sim}67\%$.
This confirms that \emph{scale drift} is the dominant mechanism of cumulative failure when many layers are perturbed: rank distortions that also shift the reconstructed weight distribution propagate destructively through the RMSNorm layers, while distortions that preserve that distribution are far more benign.

Crucially, however, the two degradation mechanisms --- scale drift and rank distortion --- differ in their correctability. Scale drift can be substantially delayed by affine correction, whereas rank distortion, once introduced, has no known post-hoc remedy (Section~\ref{sec:ablation_monotone}). The greater robustness of larger models might stem from the fact that their layers are wider and residual streams deeper, which could provide more corrective capacity to absorb per-layer perturbations. In terms of the RMSNorm absorption argument (Section~\ref{sec:ablation_sweep}), RMSNorm rescales each block's activations and so absorbs the multiplicative (variance) component of a per-layer drift, but because it does \emph{not} subtract the mean, the additive (offset) component is left to accumulate through the residual stream. Affine correction delays collapse so dramatically precisely because it removes this first-order distributional mismatch---both scale and offset---before it accumulates. The residual collapse in the corrected curves reflects higher-order shape distortions --- changes in skewness, kurtosis, and tail structure introduced by the nonlinear centroid replacement --- that an affine correction cannot remove, and which compound multiplicatively through depth.

\subsection{Centroids have a near-normal distribution}
\label{sec:normal_centroids}

For a typical MLP gate projection in the pre-compressed model ($7168\times 2816 = 20.2$M values, $K=32$), centroids form a roughly Gaussian distribution centred near zero.
Thus, the assignment of weights to clusters is highly \emph{imbalanced} --- 19 of the 32 centroids cover 90\% of weights, with the two or three near-zero centroids each assigned to over 1.2M weights.
Only a handful of ``tail'' centroids handle extreme values.
This imbalance helps explain why monotone replacement works at mid-network layers: the model's behaviour is dominated by the densely-populated near-zero centroids, and as long as the most extreme centroids remain the most extreme (in both directions), each weight position receives a qualitatively correct signal.

Recent work has identified individual ``superweights'' whose removal alone destroys model function~\citep{superweight2024}, which may seem at odds with the finding that exact magnitudes are dispensable.
However, what makes a weight critical need not be its raw magnitude but its \emph{rank} --- being the strongest or most extreme connection.
If superweights are present, clustering tends to isolate them into sparsely-populated tail clusters, preserving their rank amongst the weights.

\paragraph{Why might this hold? Three working hypotheses.} We offer three testable conjectures for why rank and variance are load-bearing. \emph{(H1) Computation is largely ordinal.} A rank-preserving affine weight map leaves the ordering of a layer's output activations unchanged (Eq.~\ref{eq:affine_output}), and the downstream SiLU and softmax are monotone, so ``which neuron/token/logit wins'' is preserved---a near-symmetry if the network's function is chiefly ordinal. \emph{(H2) Normalization fixes the scale, so variance is a second gauge.} RMSNorm absorbs a global rescaling of the weights but is calibrated to a particular activation variance; distorting the weight variance drives post-norm activations out of their trained regime, which is why variance (not distance) predicts collapse (Section~\ref{sec:controlled_grid}) and compounds with depth (Section~\ref{sec:drift}). \emph{(H3) Rank is an approximate sufficient statistic.} With most mass in a few near-zero clusters and only a handful of tail clusters carrying the extremes, \emph{which} weights are extreme (and in which direction) may summarize the layer's contribution---consistent with superweights and with why monotone quantizers (GPTQ/AWQ) preserve accuracy (Section~\ref{sec:discussion_ptq}). Appendix~\ref{app:hypotheses} turns each of these into a prediction that can fail and tests them, both in small Transformers trained from scratch and in two pretrained models that we cluster ourselves --- in each case contrasting RMSNorm, LayerNorm and un-normalized networks, which is where H2 makes an architecture-dependent prediction.
\paragraph{Scope of these hypotheses.}
We delimit these hypotheses precisely rather than leave their standing implicit. The normalisation identities of Appendix~\ref{app:activations} are exact and establish the scale clause of H2 deductively; what they do not constrain is order, so H1 and H3 rest on measurement. We therefore design the tests of Appendix~\ref{app:hypotheses} to break each hypothesis rather than to confirm it: one of them does break the offset clause of H2, and we report it and narrow the claim accordingly. Appendix~\ref{app:activations} sets out, in the same spirit, what a deductive account of the ordinal effect would require, why the loss bound available to us is uninformative, and which metric account of the effect we can rule out. The rank dependence is thus an established empirical regularity with a deliberately delimited theoretical account, and connecting it to rate--distortion characterisations of optimal scalar quantisers is the natural next step it opens.


\begin{table*}[t]
\centering
\caption{Representative compression approaches and their execution paths.}
\label{tab:method_comparison}
\small
\setlength{\tabcolsep}{4pt}
\begin{tabular}{llll}
\toprule
\textbf{Method} & \textbf{Representation} & \textbf{Optimisation} & \textbf{Runtime path} \\
\midrule
GPTQ, AWQ & Uniform / low-bit & Hessian / activ.-aware & Dense GEMM \\
AQLM & Additive codebooks & MRF / beam search & Dequant+GEMM or fused \\
VPTQ, GANQ & Non-uniform / LUT & Vector / MIQP-style & Rebuild or fused \\
\textbf{Ours} & Scalar $K$-means & $K$-means (+\,healing) & Rebuild-to-dense \\
\bottomrule
\end{tabular}
\end{table*}

\section{Relation to Quantization and Codebook Methods}
\label{sec:discussion_ptq}

We stress at the outset that this work is an \emph{analysis tool}, not a deployment-ready compressor, and the comparison below is for context rather than a competitiveness claim. Under the recommended rebuild-to-dense execution path our clustered models save \emph{disk} storage only: dense weights are reconstructed at load time, so GPU memory and runtime match the dense baseline (Table~\ref{tab:accuracy_storage}, Appendix~\ref{app:speed}; internal-harness throughput measurements confirming this are reported in Appendix~\ref{app:results_extra}). The indexed-LUT path that would reduce memory is $11$--$14\times$ slower with plain scalar $K$-means and row-level partitioning (Appendix~\ref{app:speed}). Methods such as GPTQ, AWQ, GGUF, and AQLM are mature deployment methods; we include them only to situate the accuracy--storage operating point of clustering.

Post-training quantization (PTQ) methods such as GPTQ~\citep{frantar2023gptq}, AWQ~\citep{lin2023awq}, and INT4~\citep{jacob2018quantization} also preserve weight ranks, because their quantization grids are monotone: a weight larger than another before quantization will be at least as large after. Thus, these methods implicitly preserve rank structure, and the rank dependence we observe is likely not specific to $K$-means clustering but shared by any bin-based compression scheme.

Weight clustering provides a cleaner setting for studying this dependence than PTQ for two reasons.
First, it stores the cluster labels $L$ and centroid values $c$ as explicit separate objects, so either can be perturbed independently; PTQ does not readily admit this separation. Second, $K$-means clustering places clusters where the data are densest (near-zero weights), while uniform PTQ grids must span the full range and suffer an outlier penalty; $K$-means is unaffected because outliers form their own sparsely-populated tail groups.
Indeed, randomly permuting the centroid values while keeping cluster assignments fixed --- the most extreme form of rank-breaking --- causes catastrophic degradation across all three models (see ``Rand.\ perm.'' row in Table~\ref{tab:centroid_replacement}: PPL rises to $20.67$ on Llama-Instruct, $460.2$ on the pre-compressed model, and $3{,}151$ on SmolLM2). This is even worse than rank-breaking Gaussian replacement, which at least preserves mean and variance.

Table~\ref{tab:method_comparison} places our approach among related methods.
Codebook-based methods, e.g., AQLM~\citep{egiazarian2024aqlm}, VPTQ~\citep{vptq2024}, SqueezeLLM~\citep{kim2024squeezellm}, GPTVQ~\citep{gptvq2024}, can achieve higher compression but typically require specialised kernels; train-time methods (eDKM~\citep{edkm2024}, LCD~\citep{lcd2024}) optimise during training rather than post hoc.
In contrast, our approach is applied post-training, weights are reconstructed from compact storage via Eq.~\ref{eq:reconstruction}, and no custom inference kernels are employed.

\begin{table*}[t]
\centering
\caption{
Accuracy and storage on Llama~3.1-8B-Instruct (WikiText-2 PPL, vLLM harness; see Appendix~\ref{app:eval}).
The compressed + clustered model appears as PPL~12.32 in Section~\ref{sec:ablation} and PPL~13.36 here due to different evaluation libraries (Appendix~\ref{app:eval}).}
\label{tab:accuracy_storage}
\small
\setlength{\tabcolsep}{4pt}
\begin{tabular}{l l c c c}
\toprule
\textbf{Model} & \textbf{Method} & \textbf{Disk (GB)} & \textbf{GPU (GB)} & \textbf{PPL} \\
\midrule
Llama~3.1-8B-Instruct: Baseline (FP16) & Full precision & 15.0 & 14.96 & 8.87 \\
\midrule
FP8 & Uniform quant. & 9.1 & 8.46 & 9.04 \\
AWQ INT4 & Uniform quant. & 5.7 & 5.3 & 9.35 \\
GPTQ INT4 & Uniform quant. & 5.7 & 5.3 & 9.30 \\
GGUF Q5\_K\_S & Uniform quant. & 5.3 & 5.2 & 8.99 \\
AQLM 2-bit & Codebook & 4.1 & 3.80 & 11.77 \\
\midrule
Clustered (Orig) --- no training & $K$-means & 6.9 & 14.96 & \textbf{9.32} \\
Compressed 3B-Llama & CompactifAI & 6.1 & 6.01 & 12.62 \\
Compressed + Clustered & CompactifAI + $K$-means & 3.2 & 6.01 & 13.36 \\
Compressed + Clustered + Fine-tuned & CompactifAI + $K$-means & 3.2 & 6.01 & 13.05 \\
\midrule
Compressed + Clustered + GPTQ & CompactifAI + $K$-means + INT4 & \textbf{2.6} & \textbf{2.56} & 14.21 \\
Compressed + Clustered + AWQ  & CompactifAI + $K$-means + INT4 & \textbf{2.6} & \textbf{2.55} & 13.86 \\
\bottomrule
\end{tabular}
\end{table*}

Table~\ref{tab:accuracy_storage} summarises the accuracy--storage trade-off on Llama~3.1-8B-Instruct (WikiText-2 perplexity, vLLM harness; see Appendix~\ref{app:eval}), including when quantization is added on top of compression and clustering. Weight clustering applied to the original Llama 3.1-8B-Instruct model reaches PPL~9.32 at 6.9\,GB disk without any post-compression training. This is comparable in accuracy to INT4 quantization (GPTQ/AWQ), with GPU RAM matching the dense baseline under rebuild-to-dense execution. Applying clustering on top of the pre-compressed model achieves 3.2\,GB disk ($4.7\times$ reduction from baseline), and an additional INT4 pass pushes storage to \textbf{2.6\,GB} at \textbf{2.55\,GB} GPU RAM (Pre-comp.\ + clustering + AWQ, PPL~13.86) --- the most compact configuration. Optional centroid-only fine-tuning (``healing'') recovers 30--40\% of the accuracy gap; the label structure $\mathbf{L}$ is the primary remaining bottleneck. See also Appendix~\ref{app:speed} for an execution-path analysis.

\section{Conclusions}
\label{sec:conclusions}

We have used weight clustering as a lens to study which properties of pretrained weight matrices truly matter for performance. Our central finding, demonstrated across four models of different scales and architectures (Llama~3.1-8B-Instruct, a pre-compressed Llama~3B, and SmolLM2-135M, with the single-layer rank effects further replicated on the non-Llama Qwen2.5-7B-Instruct, Appendix~\ref{app:qwen}), is that exact weight values are largely not the key factor. Instead, conditional on a fixed cluster-assignment matrix, the \emph{relative ranks} of the cluster centroids and the variance of the reconstructed weights emerge as two co-equal factors. Rank-breaking perturbations consistently cause large degradation, while rank-preserving replacements leave perplexity nearly unchanged at mid and late layers. A controlled sweep confirms this at a matched distance from the original weights, and a partial-correlation analysis isolates rank from scale, mean, and variance drift (Section~\ref{sec:controlled_grid}). We find that rank-preserving centroid edits are safe when they also preserve the mean and variance of the reconstructed weight distribution. Affine maps $f(w) = aw + b$ ($a > 0$) always preserve rank and, with $a$ and $b$ chosen to match the original mean and variance of the replacement values, preserve these moments as well; pure rescalings ($b=0$) are additionally nearly free because RMSNorm absorbs the scale. We also make explicit what rank does \emph{not} subsume: perturbing the cluster-assignment matrix $L$ itself---while leaving the centroid values untouched---degrades the model, so the spatial assignment geometry is independently load-bearing and the rank claim holds conditional on it (Section~\ref{sec:assignment_ablation}). This rank dependence is likely not unique to clustering: any bin-based compression method (including PTQ) implicitly preserves rank structure, and weight clustering is especially suited for studying this because cluster labels and representative values are stored separately. Whether the same ordinal structure governs \emph{uncompressed} models---the $K\to\infty$ limit, in which every weight is its own cluster and rank-preserving edits become arbitrary monotone reparameterizations of the full weight matrix---is a natural conjecture that our experiments do not test.

Our experiments also reveal important qualifications. Early layers are far more sensitive than mid or late layers: even rank-preserving perturbations can degrade performance substantially at layer~1, due to cumulative scale drift through the network. Progressive layer-by-layer replacement reveals that when many layers are perturbed simultaneously, \emph{scale drift} becomes the dominant collapse mechanism; however, an affine correction $w' = aw + b$ with $a > 0$ can substantially delay this drift. The two degradation mechanisms are fundamentally different in character: scale drift is correctable, while rank distortion is not --- no post-hoc remedy has been found to recover accuracy once centroid ranks are scrambled.

As a practical compression method, weight clustering applied to the original Llama~3.1-8B-Instruct achieves PPL~9.32 at 6.9\,GB disk without any fine-tuning. Combined with pre-compression and INT4 quantization, storage reaches 2.6\,GB at 2.55\,GB GPU RAM. Optional centroid-only fine-tuning recovers 30--40\% of the remaining accuracy gap at minimal cost. However, without specialised inference kernels, clustered models require reconstructing dense weight matrices at load time, limiting runtime memory savings.

\textbf{Limitations.} Our four models differ in scale, origin, tokenizer, and pretraining data, but all share the modern RMSNorm/SwiGLU/grouped-query architecture. LayerNorm-based families (e.g., GPT-2/OPT-style models), where the affine-offset mechanism of Section~\ref{sec:ablation_sweep} differs because LayerNorm centers activations, are covered only at small scale: Appendix~\ref{app:hypotheses} clusters GPT-2 124M and three byte-level models trained from scratch with RMSNorm, LayerNorm, or no normalization, and finds the rank effect in all of them. Whether it holds for a 7--8B LayerNorm model remains the most informative architectural test we cannot yet run. We also do not systematically compare base against instruction-tuned variants, nor cover larger ($>$8B), code-specialized, or strongly multilingual models. On the capability side, we probe reasoning and factual recall via zero-shot multiple-choice accuracy (Appendix~\ref{app:pertask}); we do not measure coding, instruction-following, or safety-related behavior under perturbation. We expect the rank dependence to persist wherever a monotone, bin-based reconstruction is used, but establishing this across architectures, scales, and behavioral axes is left to future work. Finally, a limitation of the contribution itself rather than of its coverage: we establish the rank and variance dependence empirically and do not derive it. The normalisation identities of Appendix~\ref{app:activations} are exact, but we have no comparable argument for why preserving centroid rank preserves function, and the hypotheses of Section~\ref{sec:normal_centroids} are tested in Appendix~\ref{app:hypotheses} rather than proved. A mechanistic account---plausibly linking the near-Gaussian centroid spectrum to rate--distortion characterisations of optimal quantisers---is the natural next step, and we regard the regularity established here as the prerequisite for it.

\textbf{Implications.} This paper contributes an understanding of clustered weights rather than a new compressor, and that understanding suggests where it could be put to use. We offer three directions, the first two speculative. \emph{Compression.} If, for fixed assignments, only the rank and the first two moments of the centroid vector need to be preserved, then centroids need not be chosen to minimise reconstruction error alone. Centroid learning could be constrained to be monotone and matched to a target mean and variance, which would shift the design effort to the assignment matrix $L$, the bottleneck that our assignment ablations (Section~\ref{sec:assignment_ablation}) and the healing results identify. \emph{Robustness and editing.} Rank-preserving, variance-matched maps form a family of nearly free reparameterisations of a clustered layer. They could serve as a safe perturbation class, for example for noise injection or for centroid fine-tuning with the ordering enforced, so that updates cannot break the structure we find to be load-bearing. \emph{Theory.} For a Gaussian source, the optimal $K$-level scalar codebook is determined by $K$, the mean, and the variance, which is the same triple our experiments find sufficient and is consistent with the near-Gaussian centroid spectrum of Section~\ref{sec:normal_centroids}. What is missing is the step from codebook optimality to function preservation in a deep network. Explaining why \emph{these} properties, and not others, govern function remains the main open question that this work raises.

Future work includes: (1)~characterising the safe region for simultaneous centroid transformations across layers; (2)~differentiable optimisation of cluster assignments to close the remaining accuracy gap without full fine-tuning (e.g., building on~\citep{edkm2024,lcd2024}); (3)~testing whether the rank dependence persists in uncompressed models and in 7--8B LayerNorm models; and (4)~formalising the conjecture that centroid rank acts as an approximate sufficient statistic for model performance, in particular whether the approximation quality can be bounded as a function of layer depth and the higher moments of the centroid transform.

\emph{Data and code availability.-} All models (Llama~3.1-8B-Instruct, SmolLM2-135M, Qwen2.5-7B-Instruct, GPT-2) and evaluation datasets (WikiText-2, ARC, HellaSwag, WinoGrande, BoolQ, MMLU) used in this work are publicly available. The code for clustering, the perturbation experiments, and evaluation is available from the corresponding author upon reasonable request.

\emph{Acknowledgements.-} We acknowledge Donostia International Physics Center (DIPC), Ikerbasque, Basque Government, Diputaci\'on de Gipuzkoa, European Innovation Council (EIC), and Spanish Government for constant support, as well as insightful discussions with the team from Multiverse Computing S.L.

\bibliographystyle{apsrev4-2}
\bibliography{references}

\appendix

\section{Additional Compression Results}
\label{app:results_extra}

\subsection{Internal Throughput (Gen TPS)}

Table~\ref{tab:speed} reports inference throughput under an internal \texttt{model.generate} harness (H200, batch=1, prompt=50, generation=50, greedy decoding).
This is a different measurement stack from the vLLM serving benchmarks in Appendix~\ref{app:speed}; the two should not be compared directly.
Weight clustering on the \emph{original} model runs at the same throughput as the FP16 baseline: under rebuild-to-dense execution, tensors are reconstructed to bf16 at load time and runtime shapes are identical.
The $\sim$3$\times$ gain for pre-compressed variants comes from the pre-compression stage reducing parameter count; clustering adds storage reduction at negligible throughput impact.

\begin{table}[t]
\centering
\caption{
Inference throughput on Llama~3.1-8B-Instruct (H200, internal harness: batch=1, prompt=50, generation=50, greedy decoding).
Gen TPS = generated tokens/second.
\emph{Not directly comparable} to vLLM serving numbers in Appendix~\ref{app:speed}.
}
\label{tab:speed}
\small
\setlength{\tabcolsep}{3pt}
\begin{tabular}{p{2.8cm} p{3.2cm} c}
\toprule
\textbf{Model} & \textbf{Method} & \textbf{Gen TPS} \\
\midrule
Baseline (FP16) & Full precision & 2194 \\
FP8 & Uniform quant. & 3316 \\
AWQ INT4 & Uniform quant. & 4606 \\
GPTQ INT4 & Uniform quant. & 1812 \\
AQLM 2-bit & Codebook & 1023 \\
\midrule
Clustering (Orig) --- no training & $K$-means (rebuild-to-dense) & $\approx$2194 \\
Pre-compressed only & Pre-compression & 6991 \\
Pre-comp.\ + clustering & Pre-comp.\ + $K$-means (rebuild-to-dense) & 6872 \\
\bottomrule
\end{tabular}
\end{table}

\section{Activation-Level Diagnostics for the Affine Argument}
\label{app:activations}

\paragraph{The two exact identities.}
Before the measurements, we state precisely the part of the argument of Section~\ref{sec:ablation_sweep} that is a derivation rather than a diagnostic. Let $y = Wx$ be the output of a linear map feeding a pre-norm block, with $\mathrm{RMSNorm}(y) = g \odot y / \sqrt{\tfrac{1}{d}\sum_i y_i^2 + \epsilon}$. \emph{(i) Scale.} Under $W \mapsto aW$ with $a>0$, both $y$ and its root-mean-square scale by $a$, so $\mathrm{RMSNorm}(ay) = \mathrm{RMSNorm}(y)$ exactly as $\epsilon \to 0$, with relative error $O(\epsilon / \lVert y \rVert^2)$ otherwise. A uniform positive rescaling of a pre-norm weight matrix is thus an exact invariance of the post-norm activations, not an approximation, which is why the affine transforms of Table~\ref{tab:monotone_sweep} with $b=0$ are nearly free. \emph{(ii) Offset.} Under $w \mapsto w + b$ applied to every entry, the output acquires $\delta = b\sum_d x_d$ in \emph{every} coordinate, i.e.\ $y \mapsto y + \delta\mathbf{1}$, a pure common mode; a common-mode shift also leaves the ordering of $y$ unchanged. LayerNorm subtracts the coordinate mean and therefore removes $\delta\mathbf{1}$ identically; RMSNorm does not, attenuating it only by the factor $\lVert y \rVert / \lVert y + \delta\mathbf{1}\rVert$ and leaving a residual that the residual stream propagates. If the offset instead lands on a pre-activation that a gate consumes before any normalisation---GELU, or the SwiGLU gate---no normalisation acts on it at all and neither identity applies, which is the asymmetry we observe in Appendix~\ref{app:hypotheses}. These statements are everything the affine argument establishes deductively; the ordinal effect itself is not derived.

\paragraph{What a full explanation would require.}
A full explanation would go beyond these identities and bound the change in loss produced by a centroid edit. We do not attempt one, and it is worth saying why. The only route available to us composes per-block Lipschitz estimates across all layers, and the product of those estimates is orders of magnitude too loose to be informative: it admits collapse in every case we measure, including the many edits that leave perplexity unchanged. One metric account can at least be ruled out rather than set aside. By the rearrangement inequality, among all assignments of a fixed multiset of replacement values to the $K$ clusters, the rank-preserving assignment minimises the weight-space displacement $\sum_k p_k (v_{\pi(k)} - c_k)^2$; rank preservation is therefore exactly the distance-minimising relabelling. That implication is correct, but it cannot be the mechanism, because displacement does not order the observed damage in the direction such an account requires: in the controlled sweep of Table~\ref{tab:controlled_grid} the partial correlation of relative $L_2$ with $\log(\mathrm{PPL}/\mathrm{PPL}_0)$ is negligible on SmolLM2 ($-0.01$) and, on Llama~3.1-8B, large but \emph{negative} ($-0.80$)---once rank and variance are held fixed, greater weight-space displacement goes with \emph{less} degradation. A first-order account in which damage grows with $\lVert \Delta W \rVert$ therefore has the wrong sign, and minimising displacement cannot be what makes rank preservation safe. What remains open is a characterisation in which rank together with two moments is \emph{sufficient} to reconstruct a layer's function; the near-Gaussian centroid spectrum of Section~\ref{sec:normal_centroids} and the rate--distortion theory of optimal scalar quantisers, in which the codebook of a Gaussian source is fixed by rank and two moments, seem the natural starting points. We flag this as the paper's main open problem rather than claiming a mechanism we have not established.

To make the normalization argument of Section~\ref{sec:ablation_sweep} concrete rather than heuristic, we measured activations directly. We confirmed that Llama~3.1 and SmolLM2 normalize with \texttt{LlamaRMSNorm}, i.e.\ RMSNorm, which divides by the activation root-mean-square but does not subtract the mean. We then applied controlled edits to a single mid layer (\texttt{layers.10.mlp.gate\_proj}) and measured the cosine similarity of the \emph{following} block's post-RMSNorm input, and of the final post-norm hidden state, against the unedited model (Table~\ref{tab:gauge}). Pure scaling is largely absorbed (post-norm cosine ${\ge}0.98$ for $w\mapsto 0.5w$ on both models); a small centred offset is nearly invisible; but a large offset of several weight standard deviations is \emph{not} absorbed (cosine drops to $0.88$ on SmolLM2 and $0.81$ on the 8B model), confirming that RMSNorm removes the multiplicative but not the additive channel of $w\mapsto aw+b$. Measured per perturbation mode, the rank-preserving modes leave post-norm activations closer to baseline (cosine $0.98$--$0.99$) than the rank-breaking modes ($0.87$--$0.94$) on both models. Finally, applying the same offset to each layer in turn and measuring the final post-norm activations shows that the earliest layer is the most sensitive (SmolLM2: cosine $0.78$ at layer~0 vs.\ ${>}0.99$ at all later layers; the trend is present but much flatter on the 8B model, minimum $0.94$), the activation-level signature of the cumulative-drift mechanism of Section~\ref{sec:ablation_multilayer}.

\begin{table*}[t]
\centering
\caption{Post-RMSNorm activation cosine similarity to the unedited model after editing a single mid layer. ``next'' is the following block's post-norm input; ``final'' is the final post-norm hidden state. $b$ is measured in units of the layer's weight standard deviation.}
\label{tab:gauge}
\small
\setlength{\tabcolsep}{5pt}
\begin{tabular}{l cc cc}
\toprule
& \multicolumn{2}{c}{\textbf{SmolLM2-135M}} & \multicolumn{2}{c}{\textbf{Llama 3.1-8B-Inst.}}\\
\cmidrule(lr){2-3}\cmidrule(lr){4-5}
\textbf{Edit / mode} & next & final & next & final \\
\midrule
Scale $a{=}0.5$              & 0.992 & 0.988 & 0.982 & 0.992 \\
Scale $a{=}2.0$              & 0.969 & 0.982 & 0.953 & 0.950 \\
Offset $b{=}{+}1\sigma$      & 0.985 & 0.992 & 0.957 & 0.975 \\
Offset $b{=}{+}5\sigma$      & 0.879 & 0.861 & 0.810 & 0.871 \\
\midrule
Rank-pres.\ (gauss.\ monotone) & 0.991 & --- & 0.980 & --- \\
Rank-pres.\ (sign-scramble)    & 1.000 & --- & 0.984 & --- \\
Rank-break (gauss.\ random)    & 0.934 & --- & 0.869 & --- \\
Rank-break (sign-pres., ord-brk) & 0.942 & --- & 0.916 & --- \\
\bottomrule
\end{tabular}
\end{table*}

\section{Attention vs.\ MLP Projection Coverage}
\label{app:coverage}

Table~\ref{tab:coverage} reports the four single-layer perturbation modes applied to every projection type at a common mid layer (layer~10). The rank-breaking modes (sign-preserving order-break; Gaussian random) are more damaging than the rank-preserving modes (order-preserving sign-scramble; Gaussian monotone) for all four attention projections and all three MLP projections, confirming the effect is not specific to \texttt{gate\_proj}. The cleanest rank-isolating contrast---Gaussian monotone vs.\ random, which share the same mean and variance---holds for \emph{every} projection in all three models. The one caveat is the order-preserving sign-scramble: although rank-preserving, it applies a centred additive offset, and where that offset is large (e.g.\ SmolLM2 \texttt{up}/\texttt{down\_proj}, which carry strongly asymmetric weight distributions) it is itself damaging---an instance of the additive-offset effect of Appendix~\ref{app:activations}, not a counterexample to the rank effect.

\begin{table}[h]
\centering
\caption{WikiText-2 PPL for the four perturbation modes across all projection types at layer~10 (Llama-3.1-8B-Instruct shown; SmolLM2 in supplementary results). kp/brk = keep/break.}
\label{tab:coverage}
\small
\setlength{\tabcolsep}{4pt}
\begin{tabular}{l cccc}
\toprule
\textbf{Projection} & \shortstack{kp\,sign\\brk\,ord} & \shortstack{kp\,ord\\brk\,sign} & \shortstack{gauss\\rand} & \shortstack{gauss\\mono} \\
\midrule
\texttt{q\_proj}    & 10.17 & 9.08 & 9.83 & 9.48 \\
\texttt{k\_proj}    & 9.86  & 9.08 & 9.77 & 9.48 \\
\texttt{v\_proj}    & 9.84  & 9.08 & 10.02 & 9.14 \\
\texttt{o\_proj}    & 10.56 & 9.38 & 10.42 & 9.18 \\
\texttt{gate\_proj} & 9.56  & 9.13 & 9.76 & 9.17 \\
\texttt{up\_proj}   & 9.63  & 9.08 & 9.79 & 9.16 \\
\texttt{down\_proj} & 9.95  & 9.08 & 9.99 & 9.18 \\
\bottomrule
\end{tabular}
\end{table}

\section{Per-Task Downstream Metrics Under Perturbation}
\label{app:pertask}

Beyond WikiText-2 perplexity, Table~\ref{tab:pertask} reports zero-shot accuracy on a suite of reasoning tasks for the clustered baseline and under single-layer centroid edits at \texttt{layers.10.mlp.gate\_proj}: a rank-preserving edit (Gaussian monotone), a rank-breaking edit (Gaussian random), and an extreme rank-breaking edit (full centroid permutation). On Llama-3.1-8B-Instruct the rank-preserving edit tracks the clustered baseline across all tasks, the rank-breaking edit degrades accuracy, and the permutation collapses it (e.g.\ ARC-Easy $0.80\!\to\!0.80\!\to\!0.75\!\to\!0.53$), showing that the rank effect extends to downstream capabilities and is not an artefact of the perplexity metric. The newly added reading-comprehension (BoolQ) and knowledge (MMLU) tasks follow the same ordering, extending the effect beyond commonsense reasoning to a distinct factual-knowledge axis: both track the baseline under the rank-preserving edit, drop under rank-breaking, and collapse under permutation. The MMLU collapse is the most striking: Llama MMLU falls from $0.682$ at baseline to $0.245$ under permutation---essentially chance (random accuracy is $0.25$)---indicating that factual knowledge is destroyed once centroid rank is broken, while a rank-preserving edit leaves it almost intact ($0.666$). The pattern is present but muted on the 135M model, which sits near chance on these tasks; HellaSwag, where it is above chance, shows the same monotone ordering ($0.43\!\to\!0.40\!\to\!0.37\!\to\!0.31$). (PIQA failed to load under our harness version and is omitted. MMLU is reported zero-shot, consistent with the rest of the table, and is omitted for the 135M model, which is at chance on it; BoolQ is likewise at chance for the 135M model and omitted there. SmolLM2 uses a 1{,}000-example subset, Llama a 500-example subset.)

\begin{table*}[t]
\centering
\caption{Zero-shot task accuracy (\texttt{lm-eval}) for the clustered model and under single-layer centroid edits at \texttt{gate\_proj}. Higher is better. ARC-c/e = ARC Challenge/Easy; perm = full centroid permutation (extreme rank-breaking).}
\label{tab:pertask}
\small
\setlength{\tabcolsep}{5pt}
\begin{tabular}{l cccccc}
\toprule
\textbf{Condition} & ARC-e & ARC-c & HellaSwag & WinoGr. & BoolQ & MMLU \\
\midrule
\multicolumn{7}{@{}l}{\textit{SmolLM2-135M}}\\
\quad Clustered baseline      & 0.469 & 0.272 & 0.425 & 0.512 & -- & -- \\
\quad Rank-preserving         & 0.461 & 0.288 & 0.398 & 0.509 & -- & -- \\
\quad Rank-breaking           & 0.458 & 0.266 & 0.365 & 0.521 & -- & -- \\
\quad Rank-breaking (perm)    & 0.303 & 0.227 & 0.313 & 0.485 & -- & -- \\
\multicolumn{7}{@{}l}{\textit{Llama-3.1-8B-Instruct}}\\
\quad Clustered baseline      & 0.804 & 0.520 & 0.704 & 0.736 & 0.814 & 0.682 \\
\quad Rank-preserving         & 0.800 & 0.528 & 0.710 & 0.742 & 0.818 & 0.666 \\
\quad Rank-breaking           & 0.750 & 0.482 & 0.682 & 0.718 & 0.776 & 0.610 \\
\quad Rank-breaking (perm)    & 0.530 & 0.318 & 0.542 & 0.600 & 0.442 & 0.245 \\
\bottomrule
\end{tabular}
\end{table*}

\section{Replication on a Non-Llama Family (Qwen2.5-7B-Instruct)}
\label{app:qwen}

To test whether the findings depend on the Llama architecture, we repeated the analysis on \textbf{Qwen2.5-7B-Instruct}, a different model family (it uses Qwen2RMSNorm, SwiGLU, and grouped-query attention, but distinct pretraining and tokenizer). We clustered all attention and MLP projections with scalar $K$-means at $K{=}64$; this raises WikiText-2 perplexity from $9.40$ to $11.83$ (internal harness). Notably, Qwen2.5 clusters \emph{less} readily than Llama ($+26\%$ vs.\ ${\sim}+5\%$ at a comparable budget), indicating that clusterability is itself architecture-dependent---which makes the replication of the \emph{rank} phenomena, below, the more informative test.

All the main effects reproduce. (i)~\textbf{Rank with distribution fixed.} Gaussian \emph{monotone} (rank-preserving) replacement is less damaging than Gaussian \emph{random} (rank-breaking) at \emph{every} projection (Table~\ref{tab:qwen_cov}). (ii)~\textbf{Rank distance.} Relabeling among rank-adjacent clusters is free ($11.83\!\to\!11.83$) while relabeling among rank-distant clusters is damaging ($11.83\!\to\!19.78$; Table~\ref{tab:assignment}). (iii)~\textbf{Downstream tasks.} A rank-preserving edit tracks the clustered baseline while rank-breaking edits degrade accuracy and a permutation collapses it, including on the added knowledge and comprehension tasks (ARC-Easy $0.77\!\to\!0.77\!\to\!0.73\!\to\!0.64$; MMLU $0.74\!\to\!0.73\!\to\!0.64\!\to\!0.47$; Table~\ref{tab:qwen_task}). (iv)~\textbf{Controlled grid.} The standardized partial correlation of centroid rank with $\log(\mathrm{PPL}/\mathrm{PPL}_0)$ is $-0.42$ at \texttt{gate\_proj} and $-0.50$ at \texttt{q\_proj}, controlling for distance and the first two moments. (v)~\textbf{Activations.} Under RMSNorm, pure scaling is absorbed (post-norm cosine $0.95$--$0.98$) but a large offset is not ($0.72$), and rank-preserving edits keep activations far closer to baseline ($0.97$) than rank-breaking ones ($0.79$--$0.82$). The additive-offset effect is in fact \emph{stronger} on Qwen: the order-preserving sign-scramble, which is rank-preserving but applies a centred offset, is highly damaging at \texttt{gate}/\texttt{up}/\texttt{down\_proj} (PPL $28$--$80$), because Qwen's per-layer centroid distributions are more asymmetric---a direct illustration of the offset channel of Appendix~\ref{app:activations}.

\begin{table*}[t]
\centering
\caption{Qwen2.5-7B-Instruct, layer~10: rank-preserving (Gaussian monotone) vs.\ rank-breaking (Gaussian random) centroid replacement, which share mean and variance and differ only in rank. The rank-preserving edit is less damaging for every projection.}
\label{tab:qwen_cov}
\small
\setlength{\tabcolsep}{4pt}
\begin{tabular}{l ccccccc}
\toprule
& \texttt{q} & \texttt{k} & \texttt{v} & \texttt{o} & \texttt{gate} & \texttt{up} & \texttt{down} \\
\midrule
Rank-pres.\ (monotone) & 13.67 & 12.19 & 12.18 & 12.12 & 12.31 & 12.20 & 12.28 \\
Rank-break (random)    & 14.15 & 12.65 & 14.04 & 13.64 & 14.11 & 14.09 & 14.12 \\
\bottomrule
\end{tabular}
\end{table*}

\begin{table*}[t]
\centering
\caption{Qwen2.5-7B-Instruct zero-shot task accuracy (\texttt{lm-eval}, 500-example subset) under single-layer centroid edits at \texttt{gate\_proj}. perm = full centroid permutation.}
\label{tab:qwen_task}
\small
\setlength{\tabcolsep}{5pt}
\begin{tabular}{l cccccc}
\toprule
\textbf{Condition} & ARC-e & ARC-c & HellaSwag & WinoGr. & BoolQ & MMLU \\
\midrule
Clustered baseline   & 0.772 & 0.548 & 0.672 & 0.694 & 0.846 & 0.739 \\
Rank-preserving      & 0.766 & 0.544 & 0.672 & 0.710 & 0.840 & 0.729 \\
Rank-breaking        & 0.728 & 0.514 & 0.620 & 0.676 & 0.806 & 0.637 \\
Rank-breaking (perm) & 0.644 & 0.416 & 0.558 & 0.516 & 0.588 & 0.471 \\
\bottomrule
\end{tabular}
\end{table*}

\section{Testing the Working Hypotheses H1--H3}
\label{app:hypotheses}

Section~\ref{sec:normal_centroids} offered three hypotheses for \emph{why} rank and variance
are the load-bearing quantities. This appendix turns each into a prediction that can fail and
tests it in two settings deliberately independent of the pipeline of
Section~\ref{sec:ablation}: small Transformers trained \emph{from scratch}, and two pretrained
models we cluster ourselves. The decisive manipulation in both is the \emph{normalization
layer}, because that is where H2 makes a falsifiable, architecture-dependent prediction. These tests are ablations, not derivations: each can falsify a hypothesis or leave it standing, but none explains the effect it probes, and we do not present them as an explanation (see Appendix~\ref{app:activations} and the Limitations of Section~\ref{sec:conclusions}).

\paragraph{What each hypothesis predicts.}
Each hypothesis is restated in the form that can fail. \emph{(H1)} Damage should track the loss
of the \emph{ordering} of a layer's output activations, not the size of the change: a large
displacement with the ordering intact should be mild, a small one that scrambles it should not.
\emph{(H2)} A \emph{model-wide} offset $w\mapsto w+b$ enters the residual stream as a
common-mode shift, so a mean-subtracting normalization should remove it and a scale-fixing one
should not, while a global rescale $w\mapsto aw$ --- absorbed by both --- should be mild under
either and fatal without a normalization; equivalently, inflating centroid variance should hurt
and restoring $(\mu,\sigma)$ at fixed rank should undo it. \emph{(H3)} Regenerating every
centroid vector as the Gaussian order statistics matched to that layer's $(\mu,\sigma)$ --- a
\emph{rank-only reconstruction} keeping only (rank, mean, variance) --- should stay near the
clustered baseline, while the same values shuffled should not.

\paragraph{Toy setup.}
We train byte-level Transformer LMs from scratch on WikiText-2 ($d=192$, $4$ layers, $6$ heads,
$d_{\text{ff}}=512$, context $128$, pre-norm blocks with SwiGLU MLPs, $1000$ steps at batch
$24$, one-cycle schedule). Three variants are identical in every respect except the
normalization layer: RMSNorm (the main-text setting), LayerNorm, and none. Each is then
clustered with the \emph{same} scalar $K$-means routine used throughout the paper ($K=32$, all
seven projections of every block) and put through the same battery of centroid edits. Perplexity
is per byte on $128$ held-out windows, so only ratios to each variant's own clustered baseline
are meaningful. Nothing here shares the paper's checkpoints, tokenizer, adaptive-$K$ rule,
healing step, or evaluation harness.

\paragraph{Pretrained setup.}
We additionally cluster two \emph{pretrained} models from their original checkpoints at a
matched $K=32$ on every projection --- SmolLM2-135M (RMSNorm) and GPT-2 124M (LayerNorm) ---
and evaluate them with the paper's own WikiText-2 harness (Appendix~\ref{app:eval}).
Interventions act on one middle-block FFN input projection unless marked otherwise, and we
record the Spearman correlation $\rho_{\mathrm{act}}$ between the perturbed and unperturbed
outputs of that layer together with the cosine similarity $\cos_{\mathrm{next}}$ of the next
block's post-normalization input, which separates ``how much the activations moved'' from
``whether their ordering survived''. No GPU was available for this round, so these tests probe
the \emph{mechanism} behind H1--H3 rather than re-establishing the 7--8B scale results of
Section~\ref{sec:ablation}.


\begin{table*}[t]
\centering
\caption{H1--H3 tested in five models: three byte-level transformers trained from scratch that are identical except for their normalization layer, and two pretrained models clustered from their original checkpoints. All are clustered with the same scalar $K$-means at $K=32$ on every projection. Entries are perplexity \emph{ratios} to each model's own clustered baseline; $\rho_{\mathrm{act}}$ is the Spearman correlation between the perturbed and unperturbed outputs of the edited layer. ``--'' marks a condition not run in that setting. Clustered baselines: 5.46 (RMSNorm), 5.51 (LayerNorm), 6.93 (No norm) byte-level, 46.48 (SmolLM2-135M, dense 26.61), 44.13 (GPT-2 124M, dense 37.37) word-level (WikiText-2, Appendix~\ref{app:eval}).}
\label{tab:hyp_all}
\footnotesize
\setlength{\tabcolsep}{3.5pt}
\begin{tabular}{llrrrrrrr}
\toprule
& & \multicolumn{3}{c}{Trained from scratch ($\times$)} & \multicolumn{2}{c}{SmolLM2-135M} & \multicolumn{2}{c}{GPT-2 124M} \\
\textbf{Intervention} & \textbf{H} & RMSNorm & LayerNorm & No norm & $\times$ & $\rho_{\mathrm{act}}$ & $\times$ & $\rho_{\mathrm{act}}$ \\
\midrule
\multicolumn{9}{l}{\emph{Edits applied to every clustered projection}} \\
Rank-only reconstruction & H3 & 2.39 & 2.48 & 7.04 & $1.8\!\times\!10^{6}$ & 0.132 & 6266 & 0.481 \\
\quad same values, order shuffled & H3 & 48.29 & 13.63 & 14.86 & $8.5\!\times\!10^{6}$ & -0.113 & $6.5\!\times\!10^{5}$ & 0.060 \\
Rank-preserving (sorted Gaussian) & H1 & 3.01 & 2.52 & 7.76 & $6.0\!\times\!10^{8}$ & 0.057 & 4403 & 0.408 \\
\quad same values, unsorted & H1 & 88.32 & 16.34 & 14.34 & -- & -- & -- & -- \\
Global rescale $a=0.5$ & H2 & 2.50 & 2.51 & 7.81 & -- & -- & -- & -- \\
Global rescale $a=2$ & H2 & 4.78 & 4.80 & $7.5\!\times\!10^{20}$ & -- & -- & -- & -- \\
Offset $b=+0.25\sigma$ & H2 & 10.46 & 1.00 & $4.2\!\times\!10^{17}$ & -- & -- & -- & -- \\
Offset $b=+0.5\sigma$ & H2 & 263.29 & 1.00 & $7.5\!\times\!10^{20}$ & -- & -- & -- & -- \\
Offset $b=+1\sigma$ & H2 & 171.08 & 1.00 & div. & -- & -- & -- & -- \\
Offset $b=+2\sigma$ & H2 & 121.03 & 1.01 & div. & -- & -- & -- & -- \\
\midrule
\multicolumn{9}{l}{\emph{Edits applied to one middle-block FFN input projection}} \\
Rank-preserving (sorted Gaussian) & H1 & 1.01 & 1.01 & 1.06 & 1.04 & 0.984 & 1.25 & 0.988 \\
\quad same values, unsorted & H1 & 1.06 & 1.05 & 1.32 & 1.50 & 0.093 & 9.43 & 0.191 \\
\quad original values, permuted & H1 & 1.24 & 1.15 & $2.8\!\times\!10^{5}$ & 112.31 & 0.039 & 172.04 & 0.045 \\
Global rescale $a=0.5$ & H2 & -- & -- & -- & 1.04 & 1.000 & 1.41 & 0.997 \\
Global rescale $a=2$ & H2 & -- & -- & -- & 1.24 & 1.000 & 1.40 & 0.999 \\
Offset $b=+0.5\sigma$ & H2 & -- & -- & -- & 1.02 & 0.919 & 1.22 & 0.998 \\
Offset $b=+1\sigma$ & H2 & -- & -- & -- & 1.08 & 0.763 & 2.07 & 0.992 \\
Offset $b=+2\sigma$ & H2 & -- & -- & -- & 1.49 & 0.515 & 28.91 & 0.968 \\
Offset $b=+4\sigma$ & H2 & -- & -- & -- & 7.98 & 0.294 & 1206 & 0.884 \\
Power $\gamma=0.5$ (variance inflated) & H2 & 1.14 & 1.21 & 29.33 & 1.34 & 0.964 & 1.98 & 0.972 \\
\quad variance restored & H2 & 1.00 & 1.00 & 1.01 & 1.03 & 0.964 & 1.04 & 0.975 \\
\midrule
\multicolumn{9}{l}{\emph{Activation-space maps, weights untouched and RMS matched}} \\
Monotone warp (rank preserved) & H1 & 1.04 & 1.04 & 1.30 & 1.02 & -- & 1.18 & -- \\
Rank-reversing map & H1 & 2.21 & 2.32 & 1.97 & 1.21 & -- & 6.76 & -- \\
Per-token shuffle (same values) & H1 & 350.62 & 59.39 & 4.31 & 1.58 & -- & 1.51 & -- \\
\bottomrule
\end{tabular}
\\[2pt]\footnotesize Toy perplexities are per byte, so only ratios are comparable across settings. Activation maps act on every FFN input projection in the toys and on the edited layer in the pretrained models. Single-layer edits are absorbed far more readily than whole-model ones, reproducing the depth-compounding effect of Section~\ref{sec:drift}.
\end{table*}

\paragraph{H1: damage tracks activation \emph{order}, not activation displacement.}
Table~\ref{tab:hyp_all} contains the pairs that make this a test
rather than an illustration. Sorting a random Gaussian centroid vector (rank preserved) and
leaving it unsorted (\emph{same values}, rank destroyed) produce comparable weight-space
displacements, but cost $\times3.0$ versus $\times88$ in the RMSNorm toy, $\times2.5$ versus
$\times16$ under LayerNorm, $\times1.04$ versus $\times1.50$ on SmolLM2-135M and $\times1.25$
versus $\times9.4$ on GPT-2, with $\rho_{\mathrm{act}}$ falling from $0.98$ to $0.09$ and
$0.19$; permuting the \emph{original} values, a smaller displacement that also scrambles the
ordering, costs $\times112$ and $\times172$. Conversely the largest activation displacement in
the toy single-layer block belongs to a rank-preserving intervention (the $\gamma{=}0.5$ power
map, relative activation $L_2$ of $3.0$) and costs only $\times1.14$. The last block of the table removes the weights from the argument altogether: a monotone warp, a rank-reversing map
and a per-token shuffle of FFN activations, each renormalized to the \emph{same} RMS, give
$\times1.04/\times2.2/\times351$ in the RMSNorm toy, $\times1.02/\times1.21/\times1.58$ on
SmolLM2 and $\times1.18/\times6.8/\times1.51$ on GPT-2 --- identical activation scale, identical
value multiset in the shuffle, only the ordering differs. The monotone map is cheapest in every
setting, so the \emph{sign} of the effect is stable; its magnitude, and even the relative
severity of reversal versus shuffling, is not (GPT-2 inverts the toys' ordering). H1 is supported
in direction and says nothing quantitative.

\paragraph{H2: normalization sets the second gauge, but only variance survives as a general
claim.}
The variance clause holds in all five settings. A global rescale $a\in\{0.5,2\}$ is
comparatively cheap wherever a normalization layer exists ($\times1.04$--$\times4.80$ across the
four normalized models) and destroys the un-normalized toy ($\times7\!\times\!10^{20}$), while
the variance-inflating $\gamma{=}0.5$ warp costs $\times1.14$--$\times29$ and restoring the
original mean and variance --- at \emph{identical} centroid rank --- returns all five to
$\times1.00$--$\times1.04$. Variance, not magnitude, is the statistic left unprotected: the
small-model counterpart of Figure~\ref{fig:weight_dist} and Section~\ref{sec:controlled_grid}.

The \emph{offset} clause holds only in the setting it was stated for. In the norm sweep, where
the normalization is the sole difference and the offset is applied to every projection, the
separation is as predicted: $\times10$ to $\times263$ under RMSNorm, free under LayerNorm
($\times1.00$--$\times1.01$), divergent with no normalization. It does \emph{not} reproduce for a
single-projection offset in the pretrained models, where the ordering reverses
($\times1.02$--$\times7.98$ on the RMSNorm model, $\times1.22$--$\times1206$ on the LayerNorm
one). The mechanism explains why rather than rescuing the prediction: an offset on an FFN
\emph{input} projection shifts a pre-activation, so the GELU or SwiGLU gate consumes it before
any normalization sees it, and a single-layer offset is therefore not a residual-stream gauge
transformation at all. The defensible statement is the narrow one: normalization fixes the
operating \emph{scale}, which makes variance the second gauge everywhere we measured; whether
the \emph{mean} is also gauge depends on where the offset enters, and is a whole-model property
that single-layer edits do not test. Offsets are also the one family of edits that damages a
model at exactly preserved centroid rank (GPT-2, $+4\sigma$: $\rho_{\mathrm{act}}=0.884$ and
$\times1206$): rank is necessary, not sufficient.

\paragraph{H3: supported at toy scale; not testable on 124M--135M models.}
Regenerating the centroids of \emph{every} projection as Gaussian order statistics matched to
each layer's $(\mu,\sigma)$ costs $\times2.4$ (RMSNorm toy) and $\times2.5$ (LayerNorm toy) ---
no worse than a generic rank-preserving monotone replacement ($\times3.0$, $\times2.5$) and far
better than the same values shuffled ($\times48$, $\times14$). On that axis rank plus two moments
carry essentially all of the usable information in $c$, which is what H3 claims. On the
pretrained models the corresponding conditions are \emph{not} informative: every whole-model
replacement collapses the model, by $\times10^{3}$--$\times10^{4}$ on GPT-2 and
$\times10^{6}$--$\times10^{8}$ on SmolLM2. The \emph{ordering} still favours H3 in both, but at
perplexities where ratios have no practical meaning, so we rest H3 on the toy sweep and on the
7--8B results of Section~\ref{sec:ablation}. We have no tested explanation for the gap: the
$\sim\!2$M toys are trained for only $1000$ steps, so their high baseline loss compresses
relative degradation, while the 124M/135M models are far less redundant than an 8B model.
Whole-model gauge freedom appears to require scale --- consistent with
Section~\ref{sec:ablation}, but not established here.

\paragraph{On setup dependence.}
Taken with the main text, the rank effect now appears in six settings that share almost no
experimental machinery --- a pre-compressed 8B Llama, Llama-3.1-8B-Instruct, Qwen2.5-7B-Instruct
(Appendix~\ref{app:qwen}), SmolLM2-135M, GPT-2, and byte-level models trained from scratch here
--- across two tokenizations, two normalization families, two evaluation harnesses, and both a
fixed and an adaptive cluster budget. In all six, rank-breaking edits cost more than
rank-preserving ones of comparable or larger magnitude, and matching $(\mu,\sigma)$ at fixed rank
recovers most of the loss. What \emph{is} setup-dependent, and now measured rather than assumed,
is the magnitude, set by how much whole-model gauge freedom scale affords, and the status of the
mean, set by where an offset enters relative to the nonlinearity. Those are the boundaries of the
claim, and we prefer to report them.

\section{Evaluation Harness Details}
\label{app:eval}

We use two evaluation stacks, which are \emph{not directly comparable} in absolute perplexity:
\begin{itemize}[leftmargin=*]
    \item \textbf{Internal harness}: WikiText-2 perplexity via EleutherAI's \texttt{lm-eval} harness~\citep{eval-harness} (\texttt{lm\_eval.evaluator.simple\_evaluate}, \texttt{batch\_size="auto"}). Used for the controlled experiments (Section~\ref{sec:ablation}).
    \item \textbf{vLLM harness}: Perplexity and throughput via vLLM~0.7.3. Used for Table~\ref{tab:accuracy_storage} (accuracy/storage) and Appendix~\ref{app:speed} (serving benchmarks).
\end{itemize}
Trends are consistent across harnesses; only within-harness comparisons are meaningful for absolute values.

\section{Inference Speed and Execution Paths}
\label{app:speed}

Weight clustering can be deployed via two execution paths: \emph{rebuild-to-dense} (weights reconstructed once at load time, then standard matrix multiplication) and \emph{indexed LUT} (direct operation on labels and centroids without materialising the dense weight matrix).
Table~\ref{tab:llama8b_config1_summary} compares both paths on the vLLM interactive serving benchmark (Llama~3.1-8B-Instruct, 512$\to$128 tokens, concurrency=1).

Throughput gains for the pre-compressed models arise from their reduced parameter count, not from weight clustering itself; clustering primarily converts this gain into additional \emph{storage} reduction, with minor throughput impact under rebuild-to-dense execution.
The LUT rows (``clustering LUT'' and ``Pre-comp.\ LUT'') show that direct indexed execution is substantially slower --- 11--14$\times$ lower throughput than rebuild-to-dense --- due to kernel bottlenecks explained below.

\begin{table*}[!t]
\centering
\caption{
vLLM interactive serving benchmark on Llama~3.1-8B-Instruct (512$\to$128 tokens, concurrency=1).
TTFT = time to first token; TPOT = time per output token.
}
\label{tab:llama8b_config1_summary}
\small
\setlength{\tabcolsep}{3pt}
\begin{tabular}{lcccccc l}
\toprule
\textbf{Model} &
\textbf{Tok/s} &
\textbf{\shortstack{TTFT\\(ms)}} &
\textbf{\shortstack{TPOT\\(ms)}} &
\textbf{\shortstack{Power\\(W)}} &
\textbf{TFLOPs} &
\textbf{\shortstack{Disk\\(GB)}} &
\textbf{Note}\\
\midrule
Original          & 155 & 25.1 & 6.30 & 266 & 0.50 & 15.0 & Baseline\\
Orig.\ clust.\ rebuild  & 157 & 25.6 & 6.21 & 232 & 0.51 & 6.9  & Same speed, less disk\\
Pre-compressed     & 259 & 18.6 & 3.74 & 188 & 0.83 & 6.1  & Best throughput\\
Pre-comp.\ clust.\ rebuild & 269 & 17.0 & 3.61 & 186 & 0.87 & 3.1 & Best trade-off\\
Orig.\ LUT          & 14  & 373  & 67.0 & 177 & 0.05 & 6.5  & Kernel-bound\\
Pre-comp.\ LUT     & 34  & 135  & 28.2 & 166 & 0.11 & 3.0  & Kernel-bound\\
\bottomrule
\end{tabular}
\end{table*}

\paragraph{Why LUT execution is slow.}
In principle, the grouped matrix-vector product can be rewritten as:
\beq
y_o = \sum_{d=1}^{D} x_d \cdot c_{L_{do}} = \sum_{k=1}^{K} c_k \cdot \underbrace{\sum_{d: L_{do}=k} x_d}_{\text{partial sum for } k}
\label{eq:lut_inference}
\eeq
reducing arithmetic from $O(DO)$ to $O(DK + KO)$ when $K \ll D$.
In practice this saving is realised only when label patterns are structured enough to allow contiguous reductions.
Table~\ref{tab:cpu_lut} shows that on CPU, all true LUT strategies are 4--6$\times$ \emph{slower} than dense computation.

\begin{table}[h]
\centering
\caption{
CPU inference time under different execution strategies (average over 3 prompts, 32 generated tokens).
``Precompute'' reconstructs weights once at load time and uses dense matrix multiplication; all other rows perform per-step LUT/gather execution.
}
\label{tab:cpu_lut}
\begin{tabular}{lcc}
\toprule
\textbf{Strategy} & \textbf{Avg.\ Time (s)} & \textbf{vs.\ Dense} \\
\midrule
Dense baseline & 0.65 & 1.00$\times$ \\
Precompute (FP32) & 0.66 & 1.02$\times$ \\
\midrule
LUT fallback & 2.94 & 4.52$\times$ slower \\
Gather-matmul & 3.01 & 4.63$\times$ slower \\
Chunked reconstruct & 4.08 & 6.28$\times$ slower \\
Vectorized scalar & 4.25 & 6.53$\times$ slower \\
\bottomrule
\end{tabular}
\end{table}

The root cause is that with row-level partitioning (chunk size $D_m=128$, varying local $K$ per chunk), the effective codebook size per output is $K_\text{eff} \approx O$: each centroid is used approximately once per output computation, eliminating the reuse that LUT acceleration depends on.
Irregular memory accesses (gather) then dominate, causing cache misses and pipeline stalls.
LUT inference can be competitive when: (i)~$K \ll O$ so that each centroid is reused many times; (ii)~label patterns have high regularity within computation tiles; and (iii)~a fused kernel keeps the centroid table in fast cache.
These conditions are exploited by methods such as AQLM~\citep{egiazarian2024aqlm} (additive codebooks) and VQ-LLM~\citep{vqllm2024} (codebook caching across the memory hierarchy), but are not met by plain scalar $K$-means with row-level partitioning.

The practical recommendation is therefore to use the rebuild-to-dense path, which is reliable, immediately deployable, and matches the throughput of the original model.

\section{Per-Layer Cluster Budget Distribution}
\label{app:k_distribution}

Table~\ref{tab:k_distribution} reports the number of MLP and attention layers assigned to each cluster budget $K$ under the adaptive per-layer selection criterion (Section~\ref{sec:method}).

\begin{table}[t]
\centering
\caption{
Number of MLP and attention projection layers assigned to each cluster budget $K$ across the three models used in this work.
}
\label{tab:k_distribution}
\small
\begin{tabular}{llcccc}
\toprule
\textbf{Model} & & $K{=}16$ & $K{=}32$ & $K{=}64$ & \textbf{Total} \\
\midrule
\multirow{2}{*}{Pre-comp.\ Llama} & MLP  & 0 & 55  & 29 & 84  \\
                                   & Attn & 0 & 91  & 21 & 112 \\
\midrule
\multirow{2}{*}{Llama 3.1-8B-Inst.} & MLP  & 0 & 78  & 18 & 96  \\
                                   & Attn & 0 & 114 & 14 & 128 \\
\midrule
\multirow{2}{*}{SmolLM2-135M}     & MLP  & 1 & 66  & 23 & 90  \\
                                   & Attn & 9 & 84  & 27 & 120 \\
\bottomrule
\end{tabular}
\end{table}

$K{=}32$ dominates across all three models.
The pre-compressed Llama variant requires more $K{=}64$ layers (50 vs.\ 32) than the original Llama~3.1-8B-Instruct, which is expected: the pre-compression stage already introduces distortion, so more centroids are needed to stay within the 0.5~PPL tolerance.
Only SmolLM2-135M has any $K{=}16$ layers (10 total, mostly in attention projections).

\end{document}